\documentclass[runningheads]{llncs}
\usepackage{makeidx}
\usepackage{graphicx}
\usepackage{amsmath,amssymb} 
\usepackage{color}

\usepackage{array}
\usepackage{booktabs}
\usepackage{colortbl}
\usepackage{hyphenat}
\usepackage{makecell}
\usepackage{multirow}
\usepackage{pgfplots}
\usepackage{subcaption}
\usepackage{tabularx}
\usepackage{tikz}
\usetikzlibrary{backgrounds,quotes,arrows.meta,calc}
\usepackage{xcolor}
\usepackage{xspace}
\usepackage{xparse}   

\newcommand{\PAR}[1]{\vskip2pt \noindent {\bf #1~}}  

\newcommand{\PARnospaceafter}[1]{\vskip4pt \noindent {\bf #1}}

\newcommand{\multibranch}{$\mathcal{M}^{\textit{MB}} $}
\newcommand{\singlebranch}{$\mathcal{M}^{\textit{SB}} $}
\newcommand{\singlebranchsplit}{$\mathcal{M}^{\textit{SB/Split}} $}
\newcommand{\stimes}{\hspace{-2pt}{\times}\hspace{-2pt}}

\def \cm{\checkmark}
\def \ph{\leavevmode\hphantom{0}}
\newcommand{\rot}[1]{\rotatebox{90}{\parbox[c][0.4cm][c]{0.7cm}{\centering #1}}}
\newcommand{\mr}[2]{\multirow{#1}{*}{#2}}
\newcommand{\task}[1]{\mr{2}{\rot{#1}}}
\newcommand{\mc}[2]{\multicolumn{#1}{c}{#2}}

\newcommand{\smalletal}{\scalebox{.8}{\etal}}
\newcolumntype{Y}{>{\centering\arraybackslash}X}

\makeatletter
\DeclareRobustCommand\onedot{\futurelet\@let@token\@onedot}
\def\@onedot{\ifx\@let@token.\else.\null\fi\xspace}

\def\etal{\emph{et al}\onedot}
\makeatother

\def \map{mAP}
\def \miou{mIoU}
\def \mioufive{mIoU\textsubscript{5}}

\definecolor{better_blue}{HTML}{4472c4}
\definecolor{better_red}{HTML}{E24A33}
\definecolor{better_green}{HTML}{8EBA42}
\definecolor{better_purple}{HTML}{D175F0}
\definecolor{awesome_orange}{HTML}{F9A91F}
\definecolor{good_gray}{HTML}{777777}
\definecolor{deep_pink}{HTML}{FF69B4}
\definecolor{shit_brown}{HTML}{8B4513}
\definecolor{random_blue}{HTML}{7DE2F0}
\definecolor{evil_green}{HTML}{366D32}

\tikzset{
  annotated cuboid/.pic={
    \tikzset{%
        every edge quotes/.append style={midway, auto},
        /cuboid/.cd,
        #1
    }
    \coordinate (orig) at (0,0,0);
    \draw [every edge/.append style={pic actions, draw=\cubfillcol}, thick, pic actions, fill=\cubfillcol]
    (0.5*\cubex,0.5*\cubey,0.5*\cubez) coordinate (o) -- ++(-\cubex,0,0) coordinate (a) -- ++(0,-\cubey,0) coordinate (b)  -- ++(\cubex,0,0) coordinate (c) -- cycle;
    \draw [every edge/.append style={pic actions, draw=\cubfillcol}, thick, pic actions, fill=\cubfillcol!75!black] (o) -- ++(0,0,-\cubez) coordinate (d) -- ++(0,-\cubey,0) coordinate (g) -- (c) -- cycle;
    \draw [every edge/.append style={pic actions, draw=\cubfillcol}, thick, pic actions, fill=\cubfillcol!75!white] (o) -- (a) -- ++(0,0,-\cubez) coordinate  (g) -- (d) -- cycle;
    \path (b)+(0,-15pt) -- (c)+(0,-15pt) node[midway] (an) {\cubeanno};
    },
    /cuboid/.search also={/tikz},
    /cuboid/.cd,
    width/.store in=\cubex,
    height/.store in=\cubey,
    depth/.store in=\cubez,
    anno/.store in=\cubeanno,
    fillcol/.store in=\cubfillcol,
    width=10,
    height=10,
    depth=10,
    anno=,
    fillcol=white,
}
\newcommand{\tensor}[9]{
    \coordinate (#1) at (#2);
    \pic [draw=#3] at (#1) {annotated cuboid={fillcol=#4, width=#5, height=#6, depth=#7,anno=#8}};
    \coordinate (#1_middle) at ($(#2)+(0.5*#5,0)$);
    \coordinate (#1_left) at ($(#2)+(-1.0*#5+#9,0)$);
}

\begin{document}
\pagestyle{headings}
\mainmatter

\title{Visual Person Understanding through\\Multi-Task and Multi-Dataset Learning}
\titlerunning{Visual Person Understanding through Multi-Task and Dataset Learning}
\authorrunning{Pfeiffer \etal}
\author{Kilian Pfeiffer, Alexander Hermans, Istv{\'a}n S{\'a}r{\'a}ndi, \\Mark Weber, and Bastian Leibe}
\institute{Visual Computing Institute, RWTH Aachen University}

\maketitle

\begin{abstract}
We address the problem of learning a single model for person re\hyp{}identification, attribute classification, body part segmentation, and pose estimation.
With predictions for these tasks we gain a more holistic understanding of persons, which is valuable for many applications.
This is a classical multi-task learning problem.
However, no dataset exists that these tasks could be jointly learned from.
Hence several datasets need to be combined during training, which in other contexts has often led to reduced performance in the past.
We extensively evaluate how the different task and datasets influence each other and how different degrees of parameter sharing between the tasks affect performance.
Our final model matches or outperforms its single-task counterparts without creating significant computational overhead, rendering it highly interesting for resource-constrained scenarios such as mobile robotics.
\end{abstract}

\section{Introduction}
Humans are arguably the most important visual category that autonomous systems need to understand in detail.
A multi-faceted understanding is especially critical for mobile robotics and autonomous driving to enable smooth human-robot interaction and pedestrian safety.
However, these are also applications with tight constraints on computational resources for reasons of cost- and energy-efficiency.
Sharing computation across tasks such as human pose estimation and attribute classification is therefore highly important.
Synergies between person-centric tasks can also emerge, potentially resulting in more accurate models.

To gain a more holistic visual understanding of a person, we jointly approach the tasks of re-identification (ReID), attribute classification, pose estimation and body part segmentation, as shown in Figure~\ref{fig:qualitative_mot}.
We argue that ReID is especially important for person understanding, as it enables tracking and merging person information across longer timespans.
Hence we place an emphasis on this task throughout our experiments.

Given the great success of deep learning in computer vision, such multi-task learning (MTL) can be realized by adding multiple output heads to a shared convolutional neural network (CNN) backbone~\cite{He17CVPR,Kendall18CVPR,Ranjan19hTPAMI}.
Several prior works have addressed some person-centric tasks jointly in this fashion.
ReID and attribute classification are known to work well together~\cite{Liu18Arxiv,Sun18ICANN,Lin17Arxiv}, similarly with pose estimation and body part segmentation~\cite{Liang18TPAMI,Papandreou18ECCV}, but so far no methods have tackled all these four tasks in a single CNN.
Possibly because no publicly available dataset has annotations for all four of these tasks.
While many interesting MTL approaches exist~\cite{Ruder17Arxiv,Kendall18CVPR}, only few of them approach multi-dataset learning, where task annotations are spread across datasets~\cite{Kokkinos17CVPR,Xiao18ECCV,Luvizon18CVPR}.
Different dataset biases make this very challenging, and it is not always possible to obtain improved results for multi-dataset learning~\cite{Kokkinos17CVPR}, as opposed to single-dataset MTL, where synergies between tasks have proven beneficial~\cite{Caruana93ICML,Kendall18CVPR}.

The design space spanned by different task-specific tricks, learning schedules, architectures, and MTL techniques is extremely large.
We therefore limit this empirical study to a single, widely used, CNN backbone with different degrees of hard parameter sharing and a simple loss weighting of the different tasks.

We first create a set of single-task baseline networks and validate their performance against state-of-the-art approaches.
We then unify these into a shared-backbone network and evaluate multi-task learning, still on a single dataset, by augmenting a ReID+attribute dataset with automatically created pose and segmentation annotations.
Finally, we evaluate how multi-dataset training affects performance.
In this setting, we find that choosing the right type of backbone normalization layers is crucial for good performance.
We consider three model variants with different degrees of parameter sharing.
Even without the use of advanced MTL techniques, our final network is able to perform all four tasks with hard parameter sharing with similar, or better performance than the baseline, rendering it very useful for practical applications.

\begin{figure}[t]
    \definecolor{mycyan}{HTML}{00A8A8}
    \definecolor{mypink}{HTML}{FF5757}
    \centering%
    \begin{subfigure}{0.495\textwidth}%
        \includegraphics[width=\textwidth]{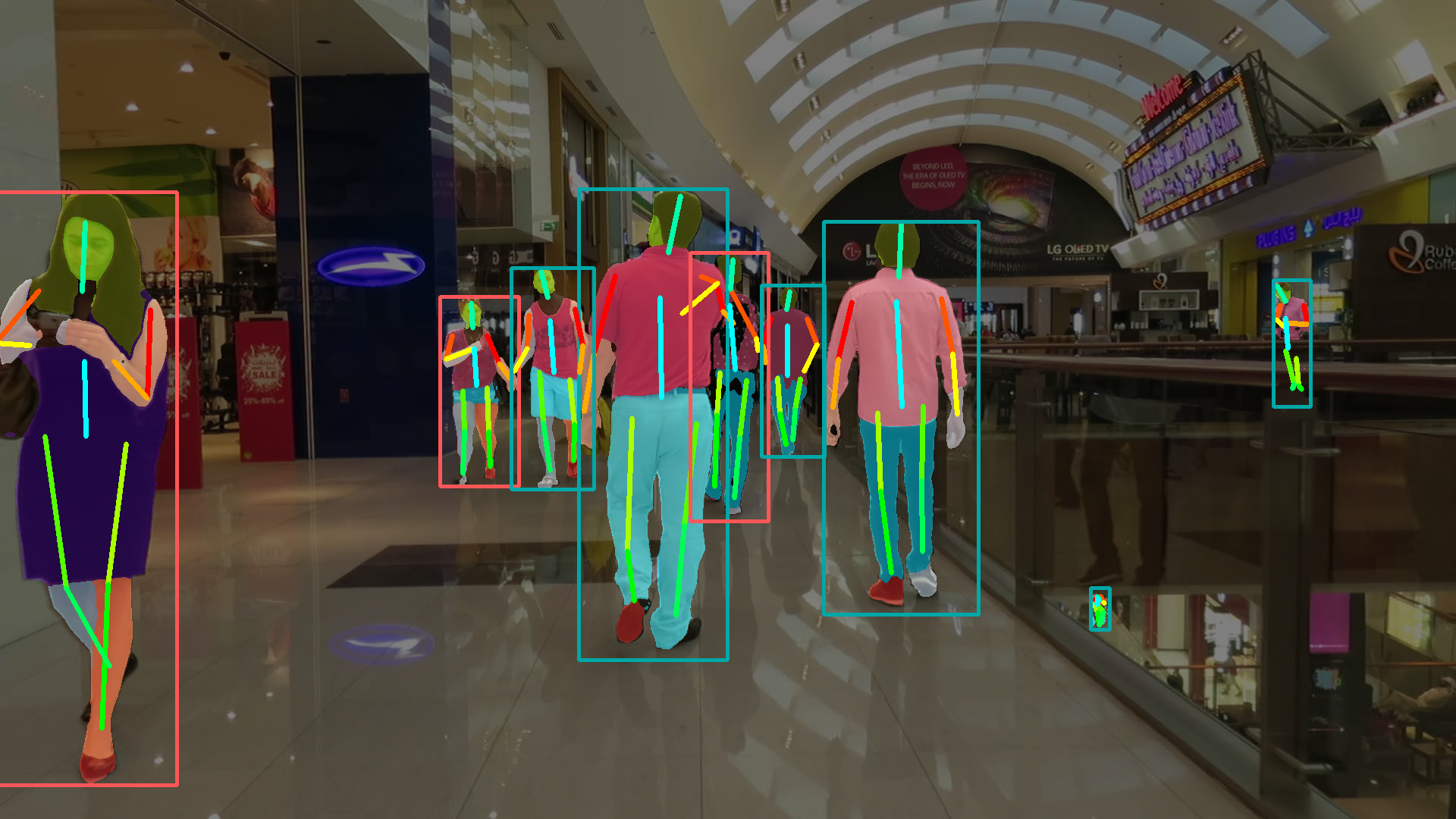}%
    \end{subfigure}\,%
    \begin{subfigure}{0.495\textwidth}%
        \includegraphics[width=\textwidth]{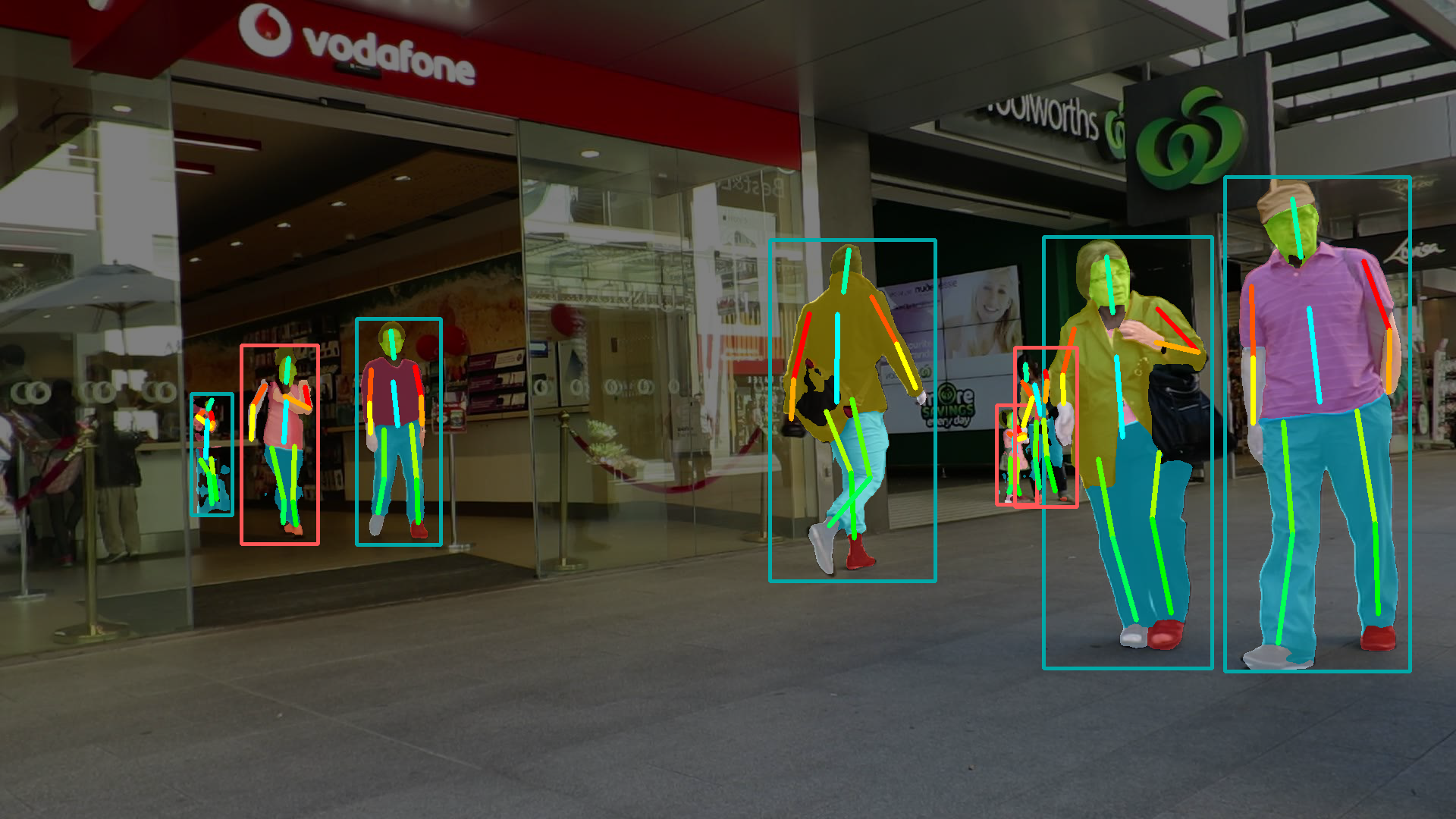}%
    \end{subfigure}%

    \begin{subfigure}{0.065\textwidth}%
        \includegraphics[width=\textwidth]{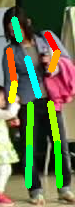}%
    \end{subfigure}\,%
    \begin{subfigure}{0.164\textwidth}%
        \includegraphics[width=\textwidth]{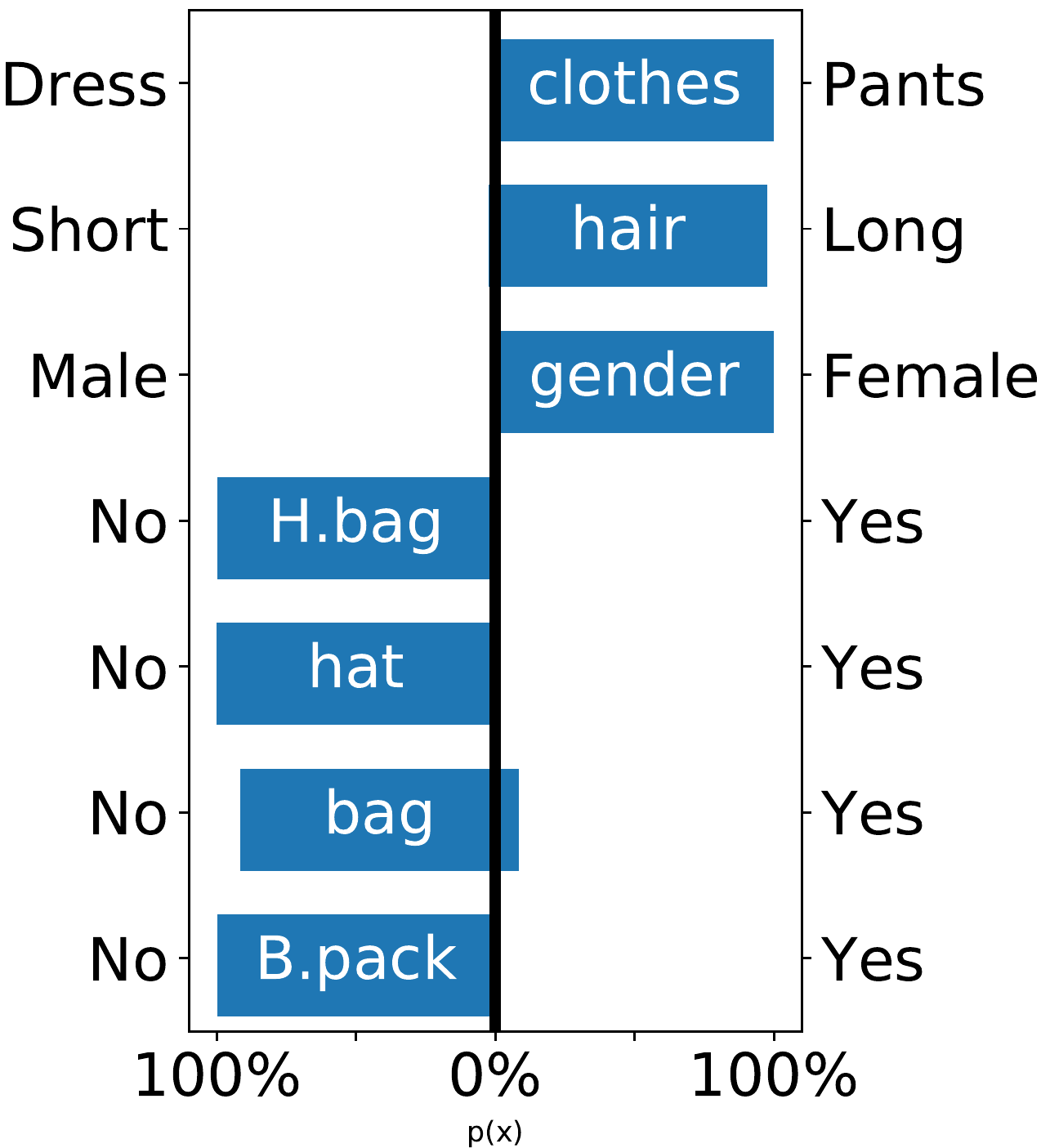}%
    \end{subfigure}~~%
    \begin{subfigure}{0.082\textwidth}%
        \includegraphics[width=\textwidth]{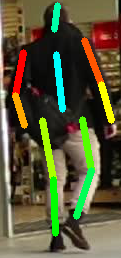}%
    \end{subfigure}\,%
    \begin{subfigure}{0.164\textwidth}%
        \includegraphics[width=\textwidth]{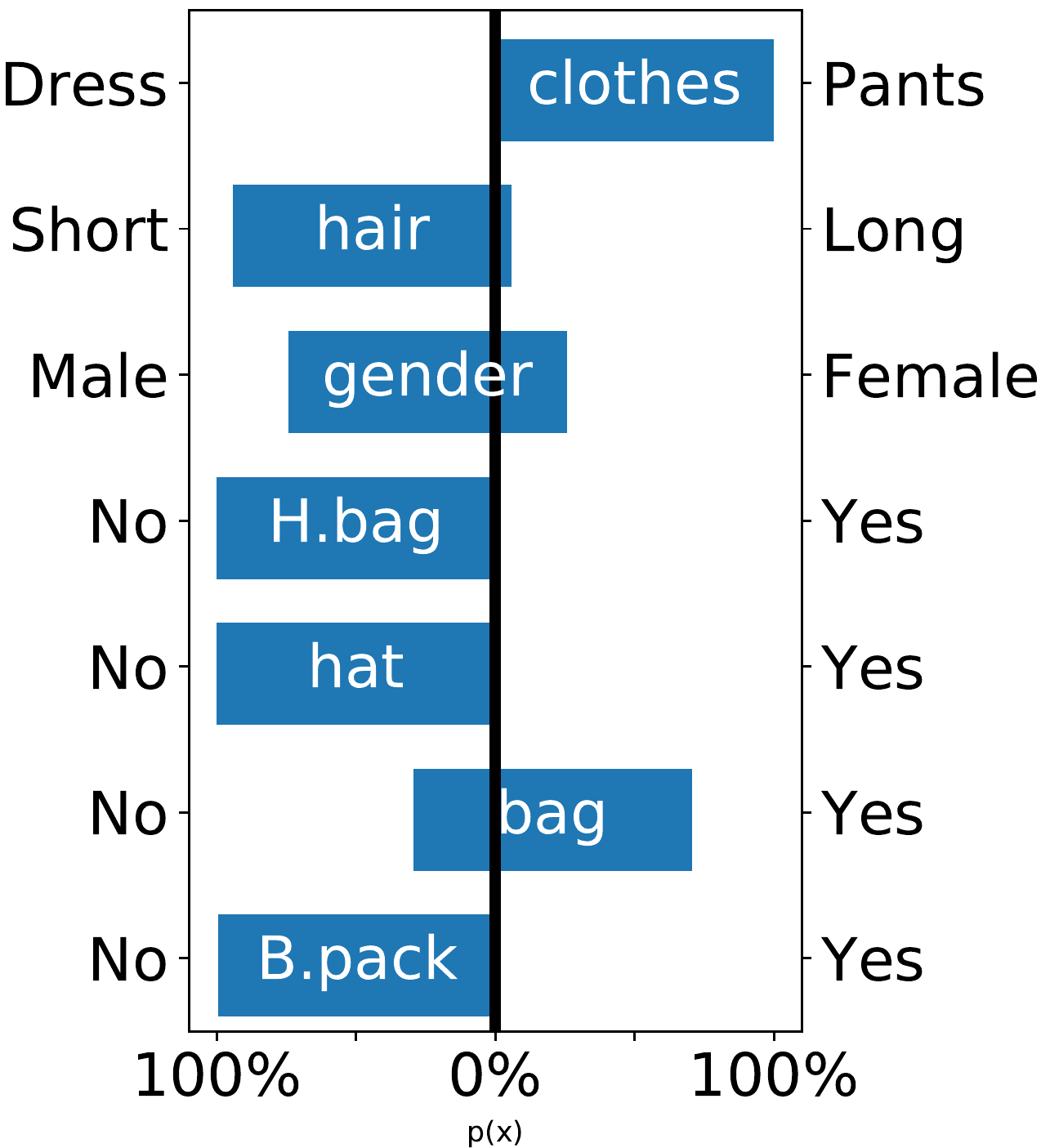}%
    \end{subfigure}~~%
    \begin{subfigure}{0.0625\textwidth}%
        \includegraphics[width=\textwidth]{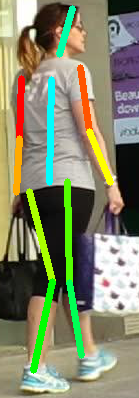}%
    \end{subfigure}\,%
    \begin{subfigure}{0.164\textwidth}%
        \includegraphics[width=\textwidth]{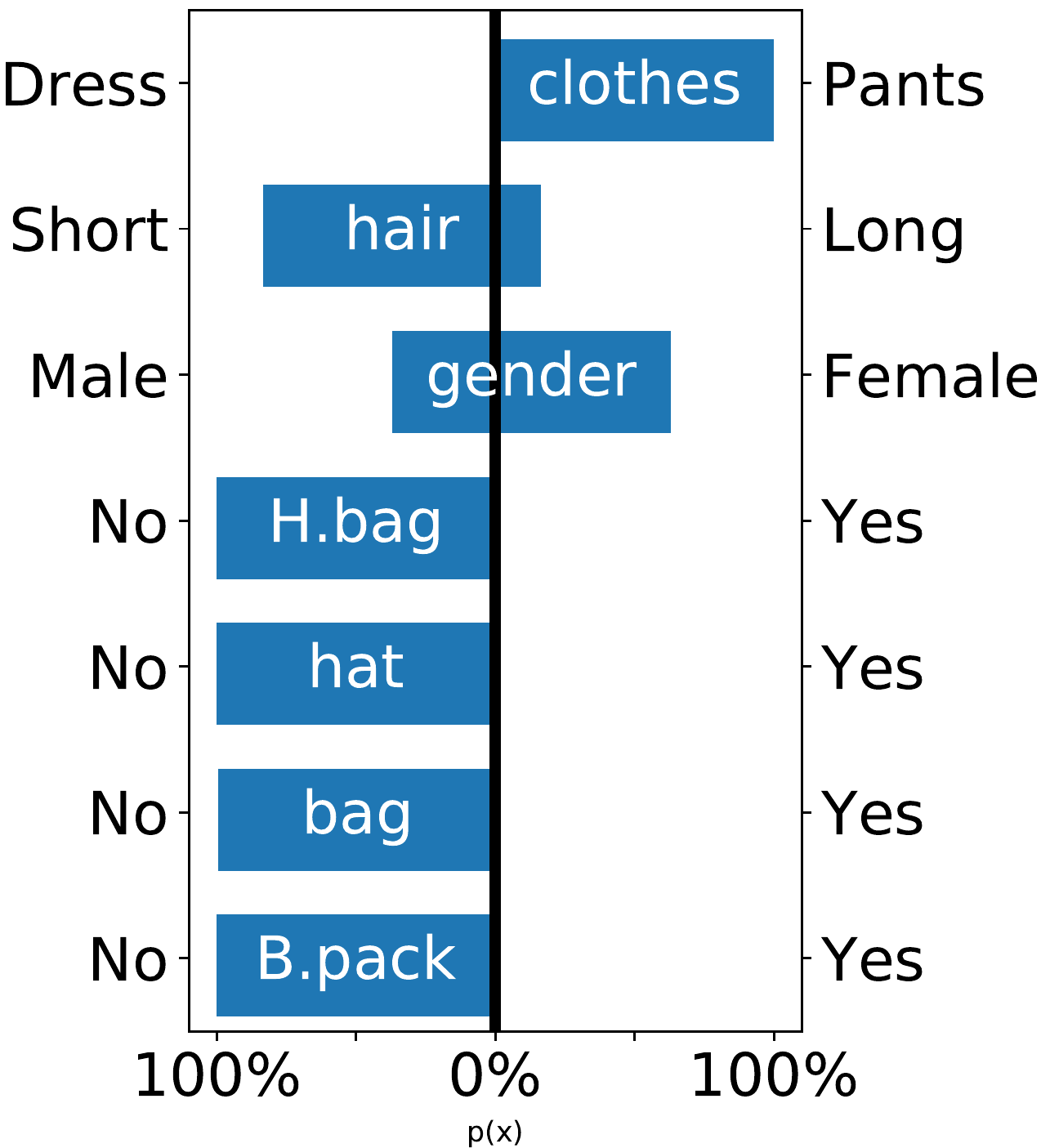}%
    \end{subfigure}~~%
    \begin{subfigure}{0.06\textwidth}%
        \includegraphics[width=\textwidth]{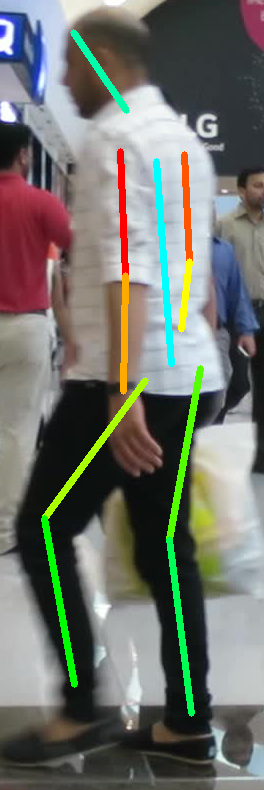}%
    \end{subfigure}\,%
    \begin{subfigure}{0.163\textwidth}%
        \includegraphics[width=\textwidth]{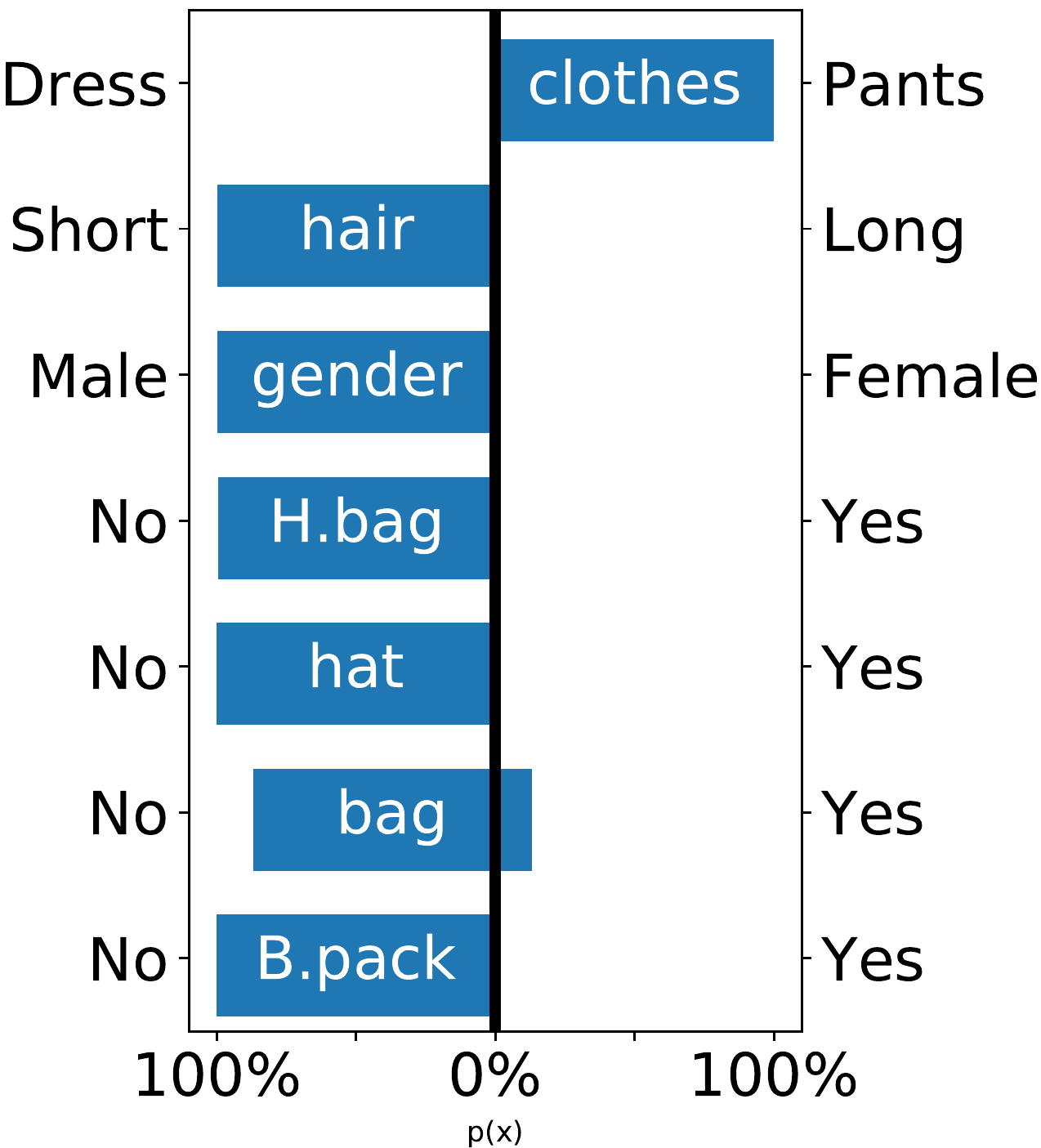}%
    \end{subfigure}%
    \caption{
        Given person detections, we perform pose estimation, body part segmentation and attribute classification jointly with person re-identification (not visualized) using a shared CNN backbone with small task-specific heads.
        Box colors correspond to gender predictions ({\color{mypink}female}, {\color{mycyan}male}).
    }
    \label{fig:qualitative_mot}
\end{figure}

\section{Related Work}
Multi-task learning (MTL) has a long history~\cite{Caruana93ICML} with the core idea that several source tasks can serve as a domain-specific inductive bias for a target task.
Ruder~\cite{Ruder17Arxiv} gives an overview of recent MTL network designs and loss or gradient merging techniques.
These developments are largely orthogonal to our experiments, since we focus on hard parameter sharing with a simple loss summation and can thus likely benefit further from some of these MTL approaches.

Several multi-task approaches exist for visual person analysis using a single dataset.
Some train pose estimation and part segmentation jointly~\cite{Papandreou18ECCV,Liang18TPAMI}.
Hyperface~\cite{Ranjan19hTPAMI} performs face detection, landmark localization, gender classification, and headpose estimation in a single network, but does not consider ReID.
He~\etal~\cite{He17CVPR} train their Mask-RCNN to jointly perform instance segmentation and human pose estimation.
Other non-person-related MTL approaches include the recently proposed panoptic segmentation~\cite{Kirillov19Arxiv}, merging instance and semantic segmentation.
Zamir~\etal~\cite{Zamir18CVPR} create a Taskonomy of indoor scene tasks, showing that many vision tasks can provide complementary supervision.

Fewer MTL approaches also learn different tasks from different datasets.
UberNet by Kokkinos~\cite{Kokkinos17CVPR} is probably the most extensive MTL approach to date, being trained on seven tasks across six different datasets.
However, they report decreased performance when trained on multiple datasets.
Xiao~\etal~\cite{Xiao18ECCV} perform several tasks on indoor scenes such as semantic, part, and texture segmentation, by pooling annotations from different datasets.
Rebuffi~\etal~\cite{Rebuffi17NIPS} introduce the visual decathlon challenge spanning ten rather different classification datasets.
They propose a shared network with domain-specific adaptation modules that modify the backbone, depending on the dataset an image comes from.
In contrast, we aim to use the same backbone on any image and perform several tasks.
To the best of our knowledge, Luvizon~\etal~\cite{Luvizon18CVPR} are the first to tackle a person-centric multi-task problem by training on multiple datasets.
They perform 2D/3D pose estimation and action recognition, but merge the tasks in a more complex way, whereas we simply attach task-specific network heads.

In the person ReID literature, MTL often refers to using several ReID losses, for example, Wang~\etal~\cite{Wang18ECCV} use a classification, triplet, and attention loss, which are all based on ReID annotations.
Some MTL approaches use attribute classification to improve ReID~\cite{Liu18Arxiv,Sun18ICANN,Lin17Arxiv}.
Other ReID approaches use pose estimation~\cite{Suh18ECCV,Qian18ECCV,Saquib18CVPR} or part segmentation~\cite{Kalayeh18CVPR}, but in the form of additional inputs, instead of producing them as outputs, like we do.

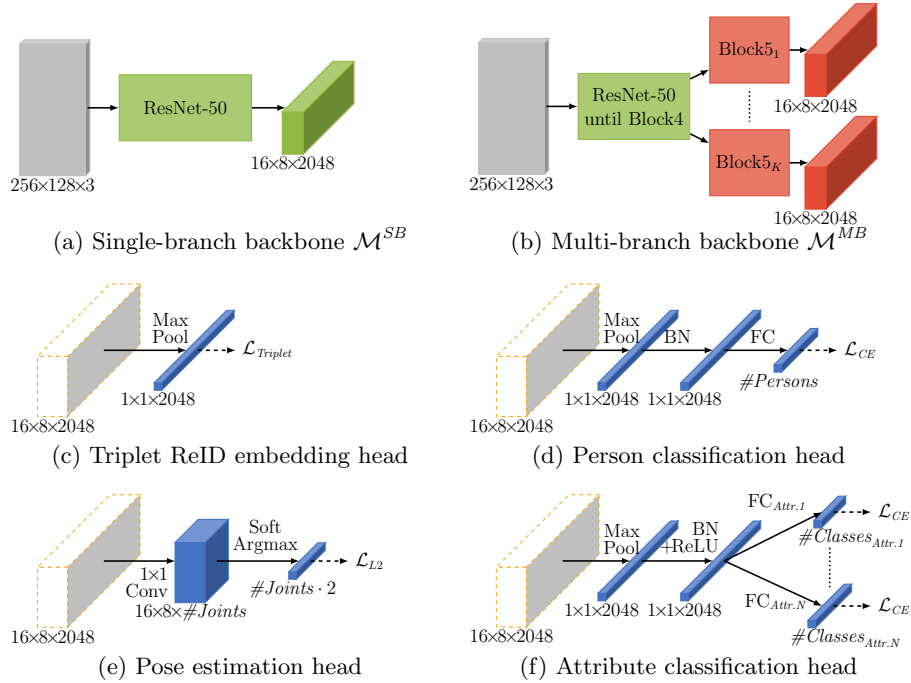
\begin{figure}[t]
    \centering
    \addtocounter{figure}{1}
    \begin{tabularx}{\textwidth}{XX}
        \vspace{-73pt}
        \scalebox{0.5}{\begin{tikzpicture}
            \fontsize{13.4}{18}\selectfont
            \tensor{image}{0,0}{lightgray}{lightgray}{1.75}{3.5}{0.3}{$256{\stimes}128{\stimes}3$}{0}
            \node[draw, thick,minimum width={100pt},minimum height={50pt}, fill=better_green!75!white, draw=better_green,anchor=west] at (1.7, 0) (net) {ResNet-50};
            \tensor{out}{7,0}{better_green}{better_green}{0.56}{1.12}{3.5}{$16{\stimes}8{\stimes}2048$}{-0.3}

            \draw[very thick, -latex]  (image_middle) -- (net);
            \draw[very thick, -latex]  (net) -- (out_left);
        \end{tikzpicture}}&
        \scalebox{0.5}{\begin{tikzpicture}
            \fontsize{13.4}{18}\selectfont
            \tensor{image}{0,0}{lightgray}{lightgray}{1.75}{3.5}{0.3}{$256{\stimes}128{\stimes}3$}{0}
            \node[draw, thick,minimum width={80pt},minimum height={50pt}, fill=better_green!75!white, draw=better_green, align=center,anchor=west] at (1.7, 0) (net) {ResNet-50 \\until Block4};
            \node[draw, thick,minimum width={60pt},minimum height={50pt}, fill=better_red!75!white, draw=better_red, align=center,anchor=west] at (5.2, 1.5) (block1) {Block5$_1$};
            \node[draw, thick,minimum width={60pt},minimum height={50pt}, fill=better_red!75!white, draw=better_red, align=center,anchor=west] at (5.2, -1.5) (block2) {Block5$_K$};
            \tensor{out1}{8.7,1.5}{better_red}{better_red}{0.56}{1.12}{3.5}{$16{\stimes}8{\stimes}2048$}{-0.3}
            \tensor{out2}{8.7,-1.5}{better_red}{better_red}{0.56}{1.12}{3.5}{$16{\stimes}8{\stimes}2048$}{-0.3}

            \draw[very thick, -latex]  (image_middle) -- (net);
            \draw[very thick, -latex]  (net) -- (block1);
            \draw[very thick, -latex]  (net) -- (block2);
            \draw[very thick, -latex]  (block1) -- (out1_left);
            \draw[very thick, -latex]  (block2) -- (out2_left);
            \draw[very thick, dotted]  (block1.south)+(0,-0.2) -- ($(block2.north)+(0,0.2)$);
        \end{tikzpicture}}\\[-5pt]
        \begin{subfigure}{\linewidth}
            \caption{Single-branch backbone \singlebranch}
            \label{fig:sb_backbone}
        \end{subfigure}&
        \begin{subfigure}{\linewidth}
            \caption{Multi-branch backbone \multibranch}
            \label{fig:mb_backbone}
        \end{subfigure}\\[7pt]
        \scalebox{0.5}{\begin{tikzpicture}
            \fontsize{13.4}{13}\selectfont
            \tensor{out}{0,0}{awesome_orange, ultra thick, dashed}{white}{0.8}{1.6}{5}{$16{\stimes}8{\stimes}2048$}{0}
            \tensor{pool}{2.8,0}{better_blue}{better_blue}{0.2}{0.2}{5}{$1\stimes1\stimes2048$}{0}

            \draw[very thick, -latex]  (out_middle) -- (pool_left) node[midway,above right,align=right] {Max\\Pool};
            \draw[very thick, dashed, -latex]  (pool_middle) -- ++(1,0) node[right] {$\mathcal{L}_{\textit{Triplet}}$};
        \end{tikzpicture}}&
        \scalebox{0.5}{\begin{tikzpicture}
            \fontsize{13.4}{13}\selectfont
            \tensor{out}{0,0}{awesome_orange, ultra thick, dashed}{white}{0.8}{1.6}{5}{$16{\stimes}8{\stimes}2048$}{0}
            \tensor{pool}{2.4,0}{better_blue}{better_blue}{0.2}{0.2}{5}{$1{\stimes}1{\stimes}2048$}{0}
            \tensor{norm}{4.6,0}{better_blue}{better_blue}{0.2}{0.2}{5}{$1{\stimes}1{\stimes}2048$}{0}
            \tensor{sm}{6.6,0}{better_blue}{better_blue}{0.2}{0.2}{2.5}{$\textit{\#Persons}$}{0}

            \draw[very thick, -latex]  (out_middle) -- (pool_left) node[midway,above right,align=right] {Max\\Pool};
            \draw[very thick, -latex]  (pool_middle) -- (norm_left) node[midway,above,align=right] {BN};
            \draw[very thick, -latex]  (norm_middle) -- (sm_left) node[pos=0.6,above] {FC};
            \draw[very thick, dashed, -latex]  (sm_middle) -- ++(1,0) node[right] {$\mathcal{L}_{\textit{CE}}$};
        \end{tikzpicture}}\\[-2pt]
        \begin{subfigure}{\linewidth}
            \caption{Triplet ReID embedding head}
            \label{fig:reid_head}
        \end{subfigure}&
        \begin{subfigure}{\linewidth}
            \caption{Person classification head}
            \label{fig:classification_head}
        \end{subfigure}\\
        \vspace{-60pt}
        \scalebox{0.5}{\begin{tikzpicture}
            \fontsize{13.4}{13}\selectfont
            \tensor{out}{0,0}{awesome_orange, ultra thick, dashed}{white}{0.8}{1.6}{5}{$16{\stimes}8{\stimes}2048$}{0}
            \tensor{conv}{3.,0}{better_blue}{better_blue}{0.8}{1.6}{1.6}{$16{\stimes}8{\stimes}\textit{\#Joints}$}{0.1}
            \tensor{argmax}{5.8,0}{better_blue}{better_blue}{0.2}{0.2}{2.0}{$\textit{\#Joints}\cdot2$}{0}

            \draw[very thick, -latex]  (out_middle) -- (conv_left) node[pos=0.6,below,align=right] {$1{\stimes}1$ \\ Conv};
            \draw[very thick, -latex]  (conv_middle) -- (argmax_left) node[pos=0.59,above,align=center] {Soft \\ Argmax};
            \draw[very thick, dashed, -latex]  (argmax_middle) -- ++(1,0) node[right] {$\mathcal{L}_{L2}$};
        \end{tikzpicture}}&
        \scalebox{0.5}{\begin{tikzpicture}
            \fontsize{13.4}{13}\selectfont
            \tensor{out}{0,0}{awesome_orange, ultra thick, dashed}{white}{0.8}{1.6}{5}{$16{\stimes}8{\stimes}2048$}{0}
            \tensor{pool}{2.4,0}{better_blue}{better_blue}{0.2}{0.2}{5}{$1{\stimes}1{\stimes}2048$}{0}
            \tensor{norm}{4.6,0}{better_blue}{better_blue}{0.2}{0.2}{5}{$1{\stimes}1{\stimes}2048$}{0}
            \tensor{sm}{7.5,1.3}{better_blue}{better_blue}{0.2}{0.2}{1.68}{\hspace{50pt}$\textit{\#Classes}_{\textit{Attr.1}}$}{0}
            \tensor{sm3}{7.5,-1.2}{better_blue}{better_blue}{0.2}{0.2}{2.5}{\hspace{50pt}$\textit{\#Classes}_{\textit{Attr.N}}$}{0}

            \draw[very thick, -latex]  (out_middle) -- (pool_left) node[midway,above right,align=right] {Max\\Pool};
            \draw[very thick, -latex]  (pool_middle) -- (norm_left) node[pos=0.65,above,align=right] {BN\\\hspace{2pt}+\hspace{-2pt}ReLU};
            \draw[very thick, -latex]  (norm_middle) -- (sm_left) node[pos=0.9,above left] {FC$_{\textit{Attr.1}}$};
            \draw[very thick, -latex]  (norm_middle) -- (sm3_left) node[pos=0.9,left] {FC$_{\textit{Attr.N}}$};
            \draw[very thick, dashed, -latex]  (sm_middle) -- ++(1,0) node[right] {$\mathcal{L}_{\textit{CE}}$};
            \draw[very thick, dashed, -latex]  (sm3_middle) -- ++(1,0) node[right] {$\mathcal{L}_{\textit{CE}}$};
            \draw[very thick, dotted]  (sm)+(0,-0.95) -- ($(sm3)+(0,0.5)$);
        \end{tikzpicture}}\\[-3pt]
        \begin{subfigure}{\linewidth}
            \caption{Pose estimation head}
            \label{fig:pose_head}
        \end{subfigure}&
        \begin{subfigure}{\linewidth}
            \caption{Attribute classification head}
            \label{fig:attribute_head}
        \end{subfigure}\\[-5pt]
    \end{tabularx}
    \caption{
        (a,b) Single and multi-branch backbones with a part shared across tasks (green), and a task-specific part (red).
        (c--f) Heads that can be attached to a branch output (shown as dashed outline).
        The segmentation head is not shown.
    }
    \label{fig:arch_backbones_heads}
\end{figure}

\section{Network Architectures}
We first introduce our single-task baseline architectures and then describe different options for merging these into multi-task architectures.
All models are built on a ResNet-50 backbone~\cite{He16CVPR}, without the global pooling and later layers.
We use a stride of one in the last block, doubling the output resolution.
The main goal while designing the baseline architectures was to make sure they are compatible amongst each other and can easily be merged to create a single multi-task architecture.

\PAR{Person Re-Identification.}
We apply global max pooling on the backbone and directly use the 2048D output as an embedding vector, in line with recently suggested best practices~\cite{Luo19CVPRW} (see Figure~\ref{fig:reid_head}).
We minimize the batch hard triplet loss~\cite{Hermans17Arxiv}, and only use horizontal flip data augmentation.

When also applying the person classification loss for ReID, we attach a BatchNorm~\cite{Ioffe15ICML} and a fully\hyp{}connected layer with softmax activation, reducing the dimensionality to the number of training persons.
We minimize the cross-entropy loss (see Figure~\ref{fig:classification_head}).

\PAR{Attribute Classification.}
On top of the ReID architecture, we add a BatchNorm layer and ReLU non-linearity, followed by separate fully\hyp{}connected layers with softmax activations for every attribute, projecting the network output down to the number of classes per attribute (see Figure~\ref{fig:attribute_head}).
The average of all attribute-specific cross-entropy losses is used, while again employing flip augmentation.

\PAR{2D Pose Estimation.}
We follow the approach by Sun~\etal~\cite{SunX18ECCV}, who predict a heatmap for every joint, from which a soft-argmax extracts 2D joint coordinates.
However, in contrast to their customized multi-stage architecture, we directly generate joint heatmaps using a $1 {\times} 1$ convolution on top of the backbone output.
We minimize the Euclidean loss between the predicted and ground-truth positions with no further heatmap-based losses (see Figure~\ref{fig:pose_head}).
Next to the random horizontal flipping, we also use rotation, translation, and scale augmentation, which is common practice for pose estimation~\cite{SunX18ECCV,Sarandi18Arxiv}.

\PAR{Body Part Segmentation.}
For part segmentation, we use the semantic segmentation branch from a recent panoptic segmentation model~\cite{Kirillov19Arxiv}.
The segmentation branch is built on top of the Feature Pyramid Network (FPN)~\cite{Lin17CVPR}, which extracts features from the backbone at several resolutions and merges them into a feature pyramid.
The levels of the pyramid are converted to quarter-resolution feature maps, which are summed and converted to a full-resolution segmentation prediction.
(See \cite{Kirillov19Arxiv} for a detailed description.)
Since our last ResNet block has a stride of one, we omit one upsampling step in the segmentation branch.
This lightweight segmentation network performs competitively on semantic segmentation benchmarks and is thus well-suited for our experiments.
Since the FPN extracts information from the backbone without changing it, it is compatible with all other baseline networks.
The segmentation is produced using a pixel-wise softmax and is trained using the bootstrapped cross-entropy loss~\cite{Wu16Arxiv} that only considers the hardest 25\% of pixels.
We apply rotation, translation and scaling augmentation during training.

\PAR{Multi-Task Architectures.}
Creating a multi-task architecture from the separate baselines described above is straightforward, since all are built on top of the same backbone.
Figure~\ref{fig:reid_head}--\ref{fig:attribute_head} show the network heads, except for the more complex segmentation head.
The simplest multi-task architecture merges all heads onto a common, single-branch backbone shown in Figure~\ref{fig:sb_backbone}.
This achieves maximal parameter sharing, since all but the final task-specific parameters are updated by every task.
Sharing all parameters might not be beneficial for all tasks, thus we also investigate a multi-branch variant, where we duplicate block 5 of our backbone for every task (Figure~\ref{fig:mb_backbone}).
To both of these backbones a set of heads can be attached, which can then be jointly trained by summing the different losses for the different tasks.
The main difference is thus the extend to which parameters and computations are shared in the backbone.
Using a single task head attached to the single-branch backbone corresponds to baseline of the respective task.

\section{Experimental Evaluation}
\subsection{Datasets}
We use three datasets to cover annotations for the four considered tasks.

\PARnospaceafter{Market-1501}~\cite{Bai17CVPR} is a ReID dataset with 32,668 images spanning 1,501 different persons.
We use the default train/test split and the single-query evaluation protocol with the standard metrics: mean Average Precision (\map) and Cumulative Matching Characteristic (CMC).
Lin~\etal~\cite{Lin17Arxiv} provide an attribute classification extension, which we evaluate using accuracy averaged over all attributes.

\PARnospaceafter{MPII}~\cite{Andriluka14CVPR} is a 2D pose estimation dataset with 15,855 well-separated person instances for training and 3,330 for validation, following the split of~\cite{Tompson15CVPR}.
The evaluation measure is the percentage of correct keypoints, which have to be closer to the ground truth than half the head size (PCKh@0.5 metric).

\PARnospaceafter{LIP}~\cite{Liang18TPAMI} provides body part segmentation and pose estimation annotations for 30,462 training and 10,000 validation images in the single-person setup.
The part segmentation consists of 20 classes, which we evaluate using the standard mean intersection over union score (\miou).
The pose annotations and evaluation are consistent with MPII.

\begin{figure}[t]
    \definecolor{lipbackground}   {HTML}{000000}
    \definecolor{liphat}          {HTML}{653700}
    \definecolor{liphair}         {HTML}{6e750e}
    \definecolor{lipglove}        {HTML}{06c2ac}
    \definecolor{lipsunglasses}   {HTML}{c79fef}
    \definecolor{lipupperclothes} {HTML}{cb416b}
    \definecolor{lipdress}        {HTML}{380282}
    \definecolor{lipcoat}         {HTML}{ceb301}
    \definecolor{lipsocks}        {HTML}{e03fd8}
    \definecolor{lippants}        {HTML}{04d9ff}
    \definecolor{lipjumpsuit}     {HTML}{11875d}
    \definecolor{lipscarf}        {HTML}{fb7d07}
    \definecolor{lipskirt}        {HTML}{ff0490}
    \definecolor{lipface}         {HTML}{89fe05}
    \definecolor{lipleftarm}      {HTML}{fcbaa0}
    \definecolor{liprightarm}     {HTML}{c5daee}
    \definecolor{lipleftleg}      {HTML}{fa6949}
    \definecolor{liprightleg}     {HTML}{6aadd5}
    \definecolor{lipleftshoe}     {HTML}{ca171c}
    \definecolor{liprightshoe}    {HTML}{8a97a3}
    \newcommand{\lipcoltableh}{%
        \begin{tikzpicture}[tight background, scale=0.75, every node/.style={font=\large}]
        \draw[white, fill=lipbackground,   draw=white] (0 * 4, 0.0) rectangle (1 * 4, 0.8) node[pos=0.5] {Background};
        \draw[white, fill=liphat,          draw=white] (1 * 4, 0.0) rectangle (2 * 4, 0.8) node[pos=0.5] {Hat};
        \draw[white, fill=liphair,         draw=white] (2 * 4, 0.0) rectangle (3 * 4, 0.8) node[pos=0.5] {Hair};
        \draw[white, fill=lipglove,        draw=white] (3 * 4, 0.0) rectangle (4 * 4, 0.8) node[pos=0.5] {Glove};
        \draw[white, fill=lipsunglasses,   draw=white] (4 * 4, 0.0) rectangle (5 * 4, 0.8) node[pos=0.5] {Sunglasses};
        \draw[white, fill=lipupperclothes, draw=white] (0 * 4,-0.8) rectangle (1 * 4, 0.0) node[pos=0.5] {Upper Clothes};
        \draw[white, fill=lipdress,        draw=white] (1 * 4,-0.8) rectangle (2 * 4, 0.0) node[pos=0.5] {Dress};
        \draw[white, fill=lipcoat,         draw=white] (2 * 4,-0.8) rectangle (3 * 4, 0.0) node[pos=0.5] {Coat};
        \draw[white, fill=lipsocks,        draw=white] (3 * 4,-0.8) rectangle (4 * 4, 0.0) node[pos=0.5] {Socks};
        \draw[black, fill=lippants,        draw=white] (4 * 4,-0.8) rectangle (5 * 4, 0.0) node[pos=0.5] {Pants};
        \draw[white, fill=lipjumpsuit,     draw=white] (0 * 4,-1.6) rectangle (1 * 4,-0.8) node[pos=0.5] {Jumpsuit};
        \draw[white, fill=lipscarf,        draw=white] (1 * 4,-1.6) rectangle (4 * 2,-0.8) node[pos=0.5] {Scarf};
        \draw[white, fill=lipskirt,        draw=white] (2 * 4,-1.6) rectangle (3 * 4,-0.8) node[pos=0.5] {Skirt};
        \draw[black, fill=lipface,         draw=white] (3 * 4,-1.6) rectangle (4 * 4,-0.8) node[pos=0.5] {Face};
        \draw[white, fill=lipleftarm,      draw=white] (4 * 4,-1.6) rectangle (5 * 4,-0.8) node[pos=0.5] {Left Arm};
        \draw[black, fill=liprightarm,     draw=white] (0 * 4,-2.4) rectangle (1 * 4,-1.6) node[pos=0.5] {Right Arm};
        \draw[white, fill=lipleftleg,      draw=white] (1 * 4,-2.4) rectangle (2 * 4,-1.6) node[pos=0.5] {Left Leg};
        \draw[white, fill=liprightleg,     draw=white] (2 * 4,-2.4) rectangle (3 * 4,-1.6) node[pos=0.5] {Right Leg};
        \draw[white, fill=lipleftshoe,     draw=white] (3 * 4,-2.4) rectangle (4 * 4,-1.6) node[pos=0.5] {Left Shoe};
        \draw[black, fill=liprightshoe,    draw=white] (4 * 4,-2.4) rectangle (5 * 4,-1.6) node[pos=0.5] {Right Shoe};
        \end{tikzpicture}%
    }
    \centering
    \setlength{\tabcolsep}{1pt}
    \renewcommand{\arraystretch}{0.5}
    \begin{tabularx}{\textwidth}{XXXXXXXX}
        \includegraphics[width=\linewidth]{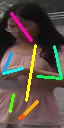}&%
        \includegraphics[width=\linewidth]{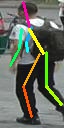}&%
        \includegraphics[width=\linewidth]{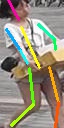}&%
        \includegraphics[width=\linewidth]{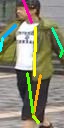}&%
        \includegraphics[width=\linewidth]{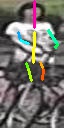}&%
        \includegraphics[width=\linewidth]{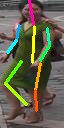}&%
        \includegraphics[width=\linewidth]{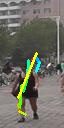}&%
        \includegraphics[width=\linewidth]{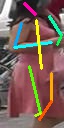}\\
        \includegraphics[width=\linewidth]{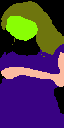}&%
        \includegraphics[width=\linewidth]{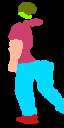}&%
        \includegraphics[width=\linewidth]{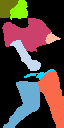}&%
        \includegraphics[width=\linewidth]{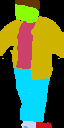}&%
        \includegraphics[width=\linewidth]{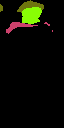}&%
        \includegraphics[width=\linewidth]{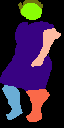}&%
        \includegraphics[width=\linewidth]{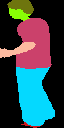}&%
        \includegraphics[width=\linewidth]{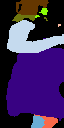}%
    \end{tabularx}
    \resizebox{\linewidth}{!}{\lipcoltableh}%
    \caption{
        Automatic annotations on the Market dataset for pose estimation (top) and part segmentation (bottom).
        Annotations quality is not always consistent, both very good (left) and rather bad (right) automatic annotations are created.
    }
    \label{fig:auto_annotations}
\end{figure}
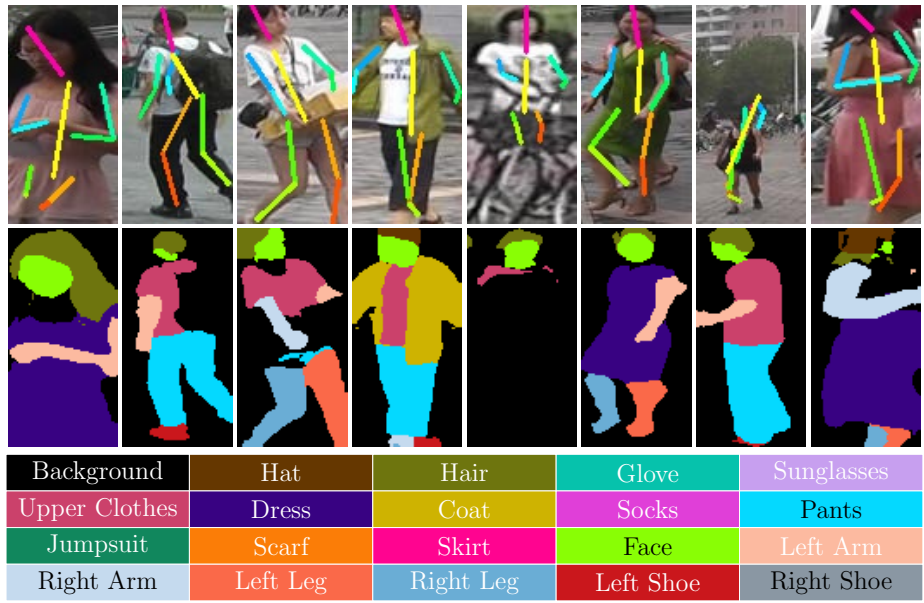

\PAR{Automatic Annotations.} To isolate the effect of multi-dataset learning, we first perform multi-task training on a single dataset.
Since no dataset provides annotations for all the tasks we consider, we extend the Market training set with automatic annotations for pose estimation and part segmentation.
To achieve this, we train baseline models on the respective datasets and use their predictions.
Visual examples are shown in Figure~\ref{fig:auto_annotations}.
Even though such annotations are of significantly lower quality than hand-annotated data, they result in interesting multi-task training results.
To reduce annotation failures, we merge several segmentation classes, similar to~\cite{Kalayeh18CVPR,Quispe18Arxiv}.
We create the background, head, upper body, lower body, and shoes classes (evaluated as \mioufive).
Since the automatic annotations are noisy, we evaluate our trained network's pose and segmentation performance on MPII and LIP instead.

\subsection{Training Setup}
We train our networks with Adam~\cite{Kingma15ICLR} using default parameters and a learning rate decay.
For multi-task training, we sum the different losses.
In multi-dataset training, we interleave mini-batches from different datasets.
We do not use mixed batches, since we require a specific batch composition for the batch hard triplet loss.
Mixed batches can also lead to noisy gradients for the task-specific parameters~\cite{Kokkinos17CVPR,Xiao18ECCV}.
We sample batches from the datasets with a frequency proportional to their sizes.
In initial experiments, GroupNorm~\cite{Wu18ECCV} performed significantly better for multi-task learning than BatchNorm~\cite{Ioffe15ICML}, hence we apply it by default unless mentioned otherwise (see Section~\ref{sec:bn_vs_gn}).
We initialize with ImageNet pretrained weights from \cite{Fu18Github}.
We use PyTorch~\cite{Paszke17NIPSWorkshop} and a Tesla V100 GPU with 16\,GB memory.
Especially training the multi-branch models require this amount of memory, whereas some of the single-task baselines can be trained with a fraction of the available GPU memory.

\begin{table}[!h]
   \centering
   \fontsize{7.6}{9}\selectfont
   \caption{Baseline person ReID performance on Market-1501.}
   \label{tab:soa_reid_market}
   \begin{tabularx}{\textwidth}{p{3.3cm}p{2.2cm}Yp{0.2cm}YYY}
        \toprule
                                        & Additional info & mAP && rank-1 & rank-5 & rank-10 \\
        \midrule
        PSE~\cite{Saquib18CVPR}         & + Pose  & 69.0 && 87.7 & 94.5 & 96.8 \\
        TriNet~\cite{Hermans17Arxiv}    & & 69.1 && 84.9 & 94.2 &   -- \\
        PN-GAN~\cite{Qian18ECCV}        & + Pose & 72.6 && 89.4 &    -- &  -- \\
        MANCS~\cite{Wang18ECCV}         & + Multi-loss & 82.3 && 93.1 &   -- &   -- \\
        SPReID~\cite{Kalayeh18CVPR}     & + Seg. & 83.4 && 93.7 & 97.6 & 98.4 \\
        Bag of Tricks~\cite{Luo19CVPRW} & + Multi-loss & 85.9 && 94.5 &   -- &   -- \\
        MGN~\cite{Wang18MM}             & + Multi-loss & 86.9 && 95.7 &   -- &   -- \\
        Pyramid~\cite{Zheng18Arxiv}     & + Multi-loss  & 88.2 && 95.7 & 98.4 & 99.0 \\
        \midrule
        Our baseline                    & & 77.4 && 91.1 & 96.9 & 98.1 \\
        \bottomrule
   \end{tabularx}
    \vspace*{3pt}

    \fontsize{7.6}{8}\selectfont
    \caption{
        Baseline attribute classification accuracy on Market-1501.
        *: Due to an unclear evaluation protocol the color attributes are not evaluated consistently.
    }
    \label{tab:soa_attr_market}
    \newcolumntype{Z}{>{\leavevmode\color{gray!75!black}}Y}
    \begin{tabularx}{\textwidth}{lYYYYYYYYYYp{1pt}Yp{1pt}ZZZ}
        \toprule
        & {\tiny Gender} & {\tiny Age} & {\tiny Hair} & {\tiny L.slv} & {\tiny L.low} & {\tiny S.clth} & \hspace*{-1pt}{\tiny B.pack} & {\tiny H.bag} & {\tiny Bag} & {\tiny Hat} && {\tiny Avg} && {\tiny C.up*} & {\tiny C.low*} & {\tiny Avg*} \\
        \midrule
        Sun \smalletal~\cite{Sun18ICANN} & 88.9 & 84.8 & 78.3 & 93.5 & 92.1 & 84.8 & 85.5 & 88.4 & 67.3 & 97.1 && 86.1 && 87.5 & 87.2 & 87.0\\
        APR\cite{Lin17Arxiv} & 86.5 & 87.1 & 83.7 & 93.7  & 93.3 & 91.5 & 82.8 & 89.0 & 75.1 & 97.1 && 88.0 && 73.4 &  69.9 & 85.3\\
        JCM\cite{Liu18Arxiv} & 89.7 & 87.4 & 82.5 & 93.7 & 93.3 & 89.2 & 85.2 & 86.2 & 86.9 & 97.2 && 89.1 && 92.4 & 93.1 & 89.7\\
        \midrule
        Our baseline & 92.9 & 87.0 & 89.7 & 93.6 & 94.8 & 94.6 & 88.0 & 89.4 & 79.7\ & 98.0 && 90.8 && 79.4 & 71.9 & 88.2\\
        \bottomrule
   \end{tabularx}
    \vspace*{3pt}

   \fontsize{7.6}{9}\selectfont
   \caption{Pose results on the MPII and LIP validation sets (PCKh@0.5).
   }
   \label{tab:soa_pose}
   \begin{tabularx}{\textwidth}{cp{2pt}lYYYYYYYp{2pt}Y}
        \toprule
        Dataset && Method & Head & Should. & Elbow & Wrist & Hip & Knee & Ankle && Avg \\
        \midrule
        \mr{4}{MPII} && Sun \smalletal~\cite{SunX18ECCV}  & -- & -- & -- & -- & -- & -- & -- && 87.3 \\
        && Tang \smalletal~\cite{Tang18ECCV}  & -- & -- & -- & -- & -- & -- & -- && 87.5 \\
        && Yang \smalletal~\cite{Yang17ICCV} & -- & -- & -- & -- & -- & -- & -- && 88.5 \\
        \cmidrule{3-12}
        && Our baseline & 96.3 & 94.4 & 87.7 & 82.1 & 86.8 & 80.6 & 74.9 && 86.6 \\
        \midrule
        && DeepLab~\cite{Chen18TPAMI,Liang18TPAMI} & 91.2 & 84.3 & 78.0 & 74.9 & 62.3 & 69.5 & 71.1 && 76.5 \\
        \mr{3}{LIP} && JPPNet~\cite{Liang18TPAMI} + Seg. & 93.2 & 89.3 & 84.6 & 82.2 & 69.9 & 78.0 & 77.3 && 82.5 \\
        \cmidrule{3-12}
        && Our baseline & 90.4 & 84.3 & 76.7 & 74.0 & 62.6 & 67.4 & 70.0 && 73.9  \\
        \bottomrule
   \end{tabularx}
    \vspace*{3pt}

    \centering
    \fontsize{7.6}{9}\selectfont
    \caption{Body part segmentation performance on the LIP validation set.}
    \label{tab:soa_seg_LIP}
    \begin{tabularx}{\textwidth}{lcYYY}
        \toprule
        & Backbone & Overall accuracy & Mean accuracy & Mean IoU \\
        \midrule
        DeepLab (from~\cite{Liang18TPAMI})     & ResNet-101 & 84.1 & 55.6 & 44.8 \\
        SS-JPPNet~\cite{Liang18TPAMI}          & ResNet-101 & 84.5 & 54.8 & 44.6 \\
        JPPNet~\cite{Liang18TPAMI} (with pose) & ResNet-101 & 86.4 & 62.2 & 51.4 \\
        CE2P~\cite{Ruan19AAAI}                 & ResNet-101 & 87.4 & 63.2 & 53.1 \\
        CaseNet~\cite{Jin19Arxiv}              & ResNet-101 &   -- &   -- & 54.4 \\
        \midrule
        Our baseline                           & ResNet-50\ph  & 84.6 & 59.5 & 47.8 \\
        \bottomrule
    \end{tabularx}
    \vspace*{-12pt}
\end{table}

\subsection{Quantitative Results}
Before discussing the multi-task and multi-datasets results, we first compare each single-task baseline to the state-of-the-art and confirm solid scores, on which we can base our multi-task experiments.
Finally, we discuss some additional interesting insights and qualitative results.

Table~\ref{tab:soa_reid_market} shows a selection of recent top-performing ReID methods.
Even though the current top approach achieves 88.2\% \map{} on the Market dataset, we still outperform many recent methods by a large margin.
Current top performing methods typically use a complex architecture~\cite{Wang18MM,Wang18ECCV,Zheng18Arxiv} or tricks such as larger input images and more elaborate augmentations~\cite{Luo19CVPRW}.
Our single-task baseline is essentially a simplified TriNet architecture~\cite{Hermans17Arxiv}, nevertheless, it still significantly improves the original mAP score of 69.14\% by over 8\%, yielding a solid baseline performance for person ReID.

For most attributes, our baseline obtains state-of-the-art accuracy as seen in Table~\ref{tab:soa_attr_market}.
Due to an ambiguous evaluation protocol, some authors~\cite{Sun18ICANN,Liu18Arxiv} evaluate multi-class color attributes as sets of binary attributes, resulting in incomparable scores.
Excluding these two attributes for a consistent comparison shows that our baseline outperforms the current top method by 1.6\%.

Table~\ref{tab:soa_pose} compares our pose estimation baseline on the MPII and LIP validation sets.
On the more commonly used MPII dataset, we achieve scores only 2\% behind the state-of-the-art and almost match Sun~\etal's original soft-argmax approach~\cite{SunX18ECCV}, even though they use several task-specific network modifications.
LIP seems to be a harder dataset and our baseline cannot directly compete with the state-of-the-art, which also uses part segmentation information.

Finally, Table~\ref{tab:soa_seg_LIP} compares our part segmentation to state-of-the-art methods on the LIP validation set.
Our baseline is 6.6\% \miou{} behind current top methods, however, all of those use the bigger ResNet-101 in combination with additional network modules or multi-scale information.
Nevertheless, our baseline outperforms several previous approaches with bigger backbones, rendering it a good starting point for our experiments.

\begin{table}[t]
    \caption{Multi-task learning on Market-1501 (manual and automatic annotations) using single- and multi-branch models, compared to the baseline of learning each task individually on Market.}
    \label{tab:multi_task_training}
    \centering
    \begin{tabularx}{\textwidth}{>{\centering\arraybackslash}p{15mm}>{\scriptsize}c>{\scriptsize}c>{\scriptsize}c>{\scriptsize}c>{\scriptsize}cYYYYY}
    \toprule
        &\mc{5}{Training: Market}& \mc{5}{Evaluation} \\
        \cmidrule(l{2pt}r{2pt}){2-6} \cmidrule(l{2pt}r{2pt}){7-11}
        &\mc{3}{Manual} & \mc{2}{Auto} &  \mc{2}{Market}  & MPII & \mc{2}{LIP} \\
        \cmidrule(l{2pt}r{2pt}){2-4} \cmidrule(l{2pt}r{2pt}){5-6} \cmidrule(l{2pt}r{2pt}){7-8} \cmidrule(l{2pt}r{2pt}){9-9} \cmidrule(l{2pt}r{2pt}){10-11}
        &\task{Tripl.}& \task{Clas.}& \task{Attr.}& \task{Pose} & \task{Seg.}& ReID & Attr. & Pose &Pose & Seg. \\
        &&&&&& \map & acc & PCKh & PCKh & \mioufive \\
        \midrule
        \mr{7}{\singlebranch}%
        &\cm&\cm&   &   &   & 77.8 &  --  &  --  &  --  &  --  \\
        &\cm&   &\cm&   &   & 78.0 & 87.9 &  --  &  --  &  --  \\
        &\cm&   &   &\cm&   & 77.9 &  --  & 27.7 & 22.6 &  --  \\
        &\cm&   &   &   &\cm& 77.6 &  --  &  --  &  --  & 48.5 \\
        &\cm&\cm&\cm&   &   & 77.8 & 87.7 &  --  &  --  &  --  \\
        &\cm&   &   &\cm&\cm& 78.6 &  --  & 30.8 & 22.4 & 49.6 \\
        &\cm&\cm&\cm&\cm&\cm& 79.2 & 88.0 & 28.8 & 21.3 & 47.9 \\
        \arrayrulecolor{lightgray}\midrule[0.2pt]\arrayrulecolor{black}
        \mr{7}{\multibranch}%
        &\cm&\cm&   &   &   & 76.9 &  --  &  --  &  --  &  -- \\
        &\cm&   &\cm&   &   & 77.5 & 87.7 &  --  &  --  &  -- \\
        &\cm&   &   &\cm&   & 77.4 &  --  & 34.4 & 27.0 &  -- \\
        &\cm&   &   &   &\cm& 77.7 &  --  &  --  &  --  & 47.6 \\
        &\cm&\cm&\cm&   &   & 78.2 & 87.5 &  --  &  --  &  -- \\
        &\cm&   &   &\cm&\cm& 77.7 &  --  & 40.4 & 28.4 & 47.9 \\
        &\cm&\cm&\cm&\cm&\cm& 78.2 & 87.9 & 39.7 & 28.1 & 46.7 \\
        \midrule
        \multicolumn{6}{c}{Baseline (individually trained)} & 77.4 & 88.2 & 46.9 & 29.9 &  48.7\\
        \bottomrule
    \end{tabularx}
    \vspace*{-20pt}
\end{table}

\PAR{Multi-Task Learning.} Having verified our baseline results, we now turn to the multi-task evaluation.
In a first round of experiments, we focus only on the Market-1501 dataset, employing our automatic annotations where necessary.
Table~\ref{tab:multi_task_training} shows several task combinations, each one using the single-branch model (\singlebranch) and the multi-branch model (\multibranch).
For these comparisons, we present baseline results trained on the same automatic annotations, to see where performance is gained due to MTL.

Some interesting observations can be made.
ReID improves in all cases when using the single-branch model, as well as for most task combinations in the multi-branch case, especially when using more than one additional task.
When considering only additional manual annotations as well as when only using automatic annotations, the \map{} score improves.
Also when considering all tasks jointly, the \map{} score still improves.
For the other tasks a different set of tasks typically is able to achieve slightly better scores.

Pose estimation suffers the most from the combination with other tasks, which becomes clear when considering the PCKh scores of the single and multi-branch models.
When only trained with the ReID task, scores become the worst, which makes sense since pose and ReID are largely orthogonal tasks.
Surprisingly though, the ReID score is not affected negatively by the pose task.

On the other hand, part segmentation benefits from being trained with ReID.
When jointly trained with pose estimation, the segmentation scores even surpass the baselines in the single-branch model, but not in the multi-branch model.

Overall, the single-branch model seems to amplify the effects that different tasks have on each other, both positive and negative, while the multi-branch model performs more similarly to the independently trained baselines.
Given that tasks share fewer parameters in the latter, this is not surprising.

All results were obtained using a fixed loss weight equal to 1.
Initial experiments with uncertainty based weighting by Kendall~\etal~\cite{Kendall18CVPR} gave inconsistent results.
In most cases the mAP score improved, while the segmentation performance decreased.
Especially for pose estimation the results were mixed, but none of the tasks saw consistent improvements.
We did not further investigate the weighting, but it is clear that the loss weights can be used to sway results towards one of the tasks.
Further research towards more clever weighting schemes should be performed.

\PAR{Multi-Dataset Training.} The multi-task experiments have clearly shown that MTL can especially be beneficial for ReID performance, but pose estimation and part segmentation results were not satisfactory, especially when compared to baselines trained on the respective datasets.
Hence, we turn to multi-dataset training using interleaved mini-batches.

Table~\ref{tab:multi_dataset_training} shows the results for a series of multi-task trainings with a focus on different dataset combinations.
Overall we observe similar effects to the single-dataset case discussed above.
ReID typically benefits from additional tasks, apart for some multi-branch models.
But with supervision from manual annotations, pose estimation and part segmentation now obtain scores that can match or improve the baseline scores.

Pose estimation still does not work well together with ReID or part segmentation.
However, when using both MPII and LIP annotations, we also match the baseline pose results, which was not possible with automatic annotations.

For part segmentation trained only with ReID, the single-branch model consistently worked better than the multi-branch model when using automatic data, where now the multi-branch model works better at the cost of a lower ReID score.
However, with additional pose estimation supervision, the single-branch model works better, as was the case before.
We can now also evaluate all 20 LIP classes and the results correlate well with the reduced class set scores.

The decrease in attribute classification accuracy is more noticeable than in the case of single-dataset training.
The effect of multi-dataset training differs per attribute.
Both the age and bag attribute accuracies improve, while the gender or hair accuracy significantly decrease.
These effects will need further investigation.

When jointly training all tasks, the multi-branch model requires more than 16\,GB GPU memory, highlighting a drawback of this backbone.
The single-branch model only needs marginally more memory than the baseline, even when training all tasks jointly, but performance is slightly worse than variants trained on fewer tasks.
Nevertheless, apart from the attribute classification task, this model can compete with all the baselines, often outperforming them.
This indicates that synergies between the tasks exist and can be exploited in a single model with hard parameter sharing.

\begin{table}[t]
    \caption{
        Multi-task and multi-dataset learning on all considered datasets.
        Results from our single and multi-branch model, as well as the split output model are compared to the our baseline.
    }
    \label{tab:multi_dataset_training}
    \begin{tabularx}{\textwidth}{c>{\scriptsize}c>{\scriptsize}c>{\scriptsize}c>{\scriptsize}c>{\scriptsize}c>{\scriptsize}cYYYYYY}
        \toprule
        &\mc{6}{Training}& \mc{6}{Evaluation} \\
        \cmidrule(l{2pt}r{2pt}){2-7} \cmidrule(l{2pt}r{2pt}){8-13} &\mc{3}{Market} & {\fontsize{7.6}{9}\selectfont MPII} & \mc{2}{LIP}  &
        \mc{2}{Market}   & MPII & \mc{3}{LIP} \\
        \cmidrule(l{2pt}r{3pt}){2-4} \cmidrule(l{0pt}r{0pt}){5-5} \cmidrule(l{3pt}r{2pt}){6-7}  \cmidrule(l{2pt}r{2pt}){8-9} \cmidrule(l{2pt}r{2pt}){10-10} \cmidrule(l{2pt}r{2pt}){11-13}
        &\task{Tripl.}& \task{Clas.}& \task{Attr.}& \task{Pose}& \task{Pose} & \task{Seg.} & ReID & Attr. & Pose  & Pose & \mc{2}{Segmentation}\\
        &&&&&&& \map & acc & PCKh  & PCKh & \miou & \mioufive \\
        \midrule
        \mr{5}{\singlebranch}%
        &\cm&   &   &\cm&   &   & 77.5 &  --  & 82.2 & 40.5 &  --  &  --  \\
        &\cm&   &   &   &   &\cm& 77.7 &  --  &  --  &  --  & 46.5 & 70.0 \\
        &\cm&   &   &\cm&   &\cm& 78.4 &  --  & 78.4 & 53.8 & 48.8 & 71.6 \\
        &\cm&   &   &\cm&\cm&\cm& 78.0 &  --  & 86.8 & 74.3 & 49.9 & 71.8 \\
        &\cm&\cm&\cm&\cm&\cm&\cm& 78.3 & 87.1 & 86.7 & 73.8 & 49.6 & 71.6 \\
        \arrayrulecolor{lightgray}\midrule[0.25pt]\arrayrulecolor{black}
        \mr{5}{\multibranch}%
        &\cm&   &   &\cm&   &   & 76.9 &  --  & 83.8 & 41.6 &  --  &  --  \\
        &\cm&   &   &   &   &\cm& 76.8 &  --  & --   &  --  & 47.2 & 70.8 \\
        &\cm&   &   &\cm&   &\cm& 77.8 &  --  & 83.9 & 52.9 & 47.5 & 71.0 \\
        &\cm&   &   &\cm&\cm&\cm& 77.9 &  --  & 86.9 & 75.0 & 48.5 & 71.6 \\
        &\cm&\cm&\cm&\cm&\cm&\cm& \mc{6}{ {\fontsize{6.6}{9}\selectfont ----- out of GPU memory ----- } } \\
        \arrayrulecolor{lightgray}\midrule[0.25pt]\arrayrulecolor{black}
        \mr{2}{\singlebranchsplit}%
        &\cm&   &   &\cm&   &   & 77.9 &  --  & 86.7 & 44.5 &  --  &  --  \\
        &\cm&\cm&\cm&\cm&\cm&\cm& 79.1 & 86.7 & 86.5 & 74.4 & 49.6 & 71.6 \\
        \midrule
        \mc{7}{Baselines (individually trained)} & 77.4 & 88.2  & 86.6 & 73.9 & 47.8 & 71.0 \\
        \bottomrule
    \end{tabularx}
\end{table}

\begin{figure}[!t]
    \vspace*{10pt}
    \centering
    \scalebox{0.55}{\begin{tikzpicture}
         \fontsize{13.4}{13}\selectfont
        \tensor{image}{0,0}{lightgray}{lightgray}{1.75}{3.5}{0.3}{$256{\times}128{\times}3$}{0}
        \node[draw, thick,minimum width={100pt},minimum height={50pt}, fill=better_green!75!white, draw=better_green,anchor=west] at (1.7, 0) (net) {ResNet-50};
        \tensor{out}{7,0}{better_green}{better_green}{0.56}{1.12}{3.5}{$16{{\times}}8{{\times}}2048$}{-0.3}
        \tensor{split1}{10.9,0.15}{better_green}{better_green}{0.56}{1.12}{3.0}{}{-0.3}
        \path (d)+(0,12pt) -- (g)+(0,12pt) node[midway] (an) {$16{\times}8{\times}2048-\#Joints$};
        \tensor{split2}{10.0,-0.75}{better_green}{better_green}{0.56}{1.12}{0.4}{$16{\times}8{\times}\#Joints$}{+0.2}

        \tensor{pool}{16.6,0.11}{better_blue}{better_blue}{0.2}{0.2}{3.0}{}{0}
        \path (d)+(0,12pt) -- (g)+(0,12pt) node[midway] (an) {$1{\times}1{\times}2048-\#Joints$};
        \tensor{argmax}{15.7,-0.75}{better_blue}{better_blue}{0.2}{0.2}{0.4}{$\#Joints\cdot2$}{0}

        \draw[very thick, -latex]  (image_middle) -- (net);
        \draw[very thick, -latex]  (net) -- (out_left);
        \draw[very thick, -latex]  (out_middle) -- (split1_left) node[midway,above] {Split};
        \draw[very thick, -latex]  (out_middle) -- (split2_left);
        \draw[very thick, -latex]  (split1_middle) -- (pool_left) node[midway,above] {Max Pool};
        \draw[very thick, -latex]  (split2_middle) -- (argmax_left) node[midway,below] {Soft Argmax};
        \draw[very thick, dashed, -latex]  (pool_middle) -- ++(1,0) node[right] {$\mathcal{L}_{Triplet}$};
        \draw[very thick, dashed, -latex]  (argmax_middle) -- ++(1,0) node[right] {$\mathcal{L}_{L2}$};

    \end{tikzpicture}}
    \caption{
        Single-branch model with a split-pose head (\singlebranchsplit).
        The backbone output is split channel-wise into two tensors, one for body joint heatmaps and one for other task heads.
        For simplicity only the triplet loss $\mathcal{L}_{Triplet}$ is shown.
    }
    \label{fig:sb_split}
    \vspace*{-10pt}
\end{figure}
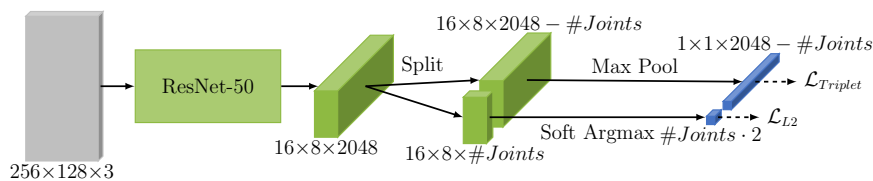

\subsection{ReID and Pose Interference}
The previous experiments suggest that the single-branch model is better overall, except that pose estimation degrades when trained jointly with ReID.
As a possible solution, we make a slight modification to our single-branch model, where we split the backbone output into two separate tensors: one for pose heatmaps and one for the remaining tasks.
This allows sharing all backbone weights, while still reducing interference between pose estimation and ReID gradients at the backbone output.
Figure~\ref{fig:sb_split} shows this variant (with only ReID as an additional task).
This modification only slightly changes most scores, but boosts the ReID performance by 0.8\% as shown in the lower block of Table~\ref{tab:multi_dataset_training}.

\begin{table}[t]
    \fontsize{7.6}{9}\selectfont
    \caption{
        Comparison of multi-task training approaches and backbone normalizations.
        The three column-blocks show, respectively, the pose results trained only on MPII, ReID and pose after additional fine-tuning for ReID on Market, and finally, when jointly training both at once.
        LIP is only used for evaluation here.
    }
    \label{tab:gn_vs_bn}
    \begin{tabularx}{\textwidth}{cp{2pt}cp{2pt}Yp{0pt}Yp{10pt}Yp{0pt}Yp{0pt}Yp{10pt}Yp{0pt}Yp{0pt}Y}
        \toprule
        &&&&\multicolumn{3}{c}{Train on MPII} &~$\rightarrow$& \multicolumn{5}{c}{Fine-tune on Market} && \multicolumn{5}{c}{Train jointly} \\
        \cmidrule{5-7} \cmidrule{9-13} \cmidrule{15-19}
        &&&& MPII && LIP && Market && MPII && LIP && Market && MPII && LIP\\
        \cmidrule{5-5} \cmidrule{7-7} \cmidrule{9-9} \cmidrule{11-11} \cmidrule{13-13} \cmidrule{15-15} \cmidrule{17-17} \cmidrule{19-19}
        &&&&           PCKh  && PCKh && \map &&  PCKh  &&PCKh  && \map && PCKh &&PCKh\\
        \midrule
        \mr{2}{\singlebranch}%
        && BatchNorm && 86.5 && 41.3 && 73.9 &&   16.7 && 10.6 && 51.5 && 79.5 && 32.8\\
        && GroupNorm && 86.4 && 43.6 && 72.6 && \ph9.0 && 10.0 && 77.5 && 82.2 && 40.5\\
        \midrule
        \mr{2}{\multibranch}%
        && BatchNorm && 86.2 && 39.5 && 77.7 &&   57.4 && 15.6 && 73.9 && 80.1 && 32.8\\
        && GroupNorm && 86.6 && 43.3 && 75.6 &&   70.1 && 33.9 && 76.9 && 83.8 && 41.6\\
        \bottomrule
    \end{tabularx}
\end{table}

\subsection{Pretraining on MPII}
As an alternative to joint training of tasks, we also evaluate pretraining the backbone on MPII and using these weights to initialize ReID training on Market.
As Table~\ref{tab:gn_vs_bn} shows, the resulting mAP scores cannot match the baseline and the ReID training additionally has a very negative impact on pose estimation, while the joint training for both tasks gains the previously discussed results.
Especially the single-branch model no longer produces a useful pose estimation after finetuning, probably because the largely pose-invariant ReID training erases pose-related information from the backbone output.

\subsection{GroupNorm vs BatchNorm}\label{sec:bn_vs_gn}
Initially we used a backbone with BatchNorm, which did not perform well during multi-dataset training.
We hypothesize that different datasets result in different batch statistics, hence the overall collected statistics do not match the per-dataset train-time statistics, ultimately leading to worse results at test-time.
While GroupNorm was initially developed to train with small mini-batches, a further benefit is that it does not accumulate batch statistics during training and thus should also improve joint training.
Table~\ref{tab:gn_vs_bn} shows a comparison of the two different normalizations in the backbone, for the example of pose estimation and person ReID.
The GroupNorm results for joint training are consistently better, outperforming the single-branch BatchNorm mAP score by over 25\%.
Furthermore, the generalization to different datasets (in this case LIP) is better for the GroupNorm models, which could again be explained by the fact that BatchNorm uses possibly unequal dataset statistics.
Given these results, we use GroupNorm in all other experiments.

Somewhat surprising are the results of the aforementioned finetuning experiment when using BatchNorm.
Here the multi-branch model even outperforms the baseline ReID performance, however, at the cost of worse pose estimation results.
For the single-branch model the BatchNorm variant also performs slightly better than the GroupNorm variant, but both ReID and pose estimation performance cannot match the baseline.

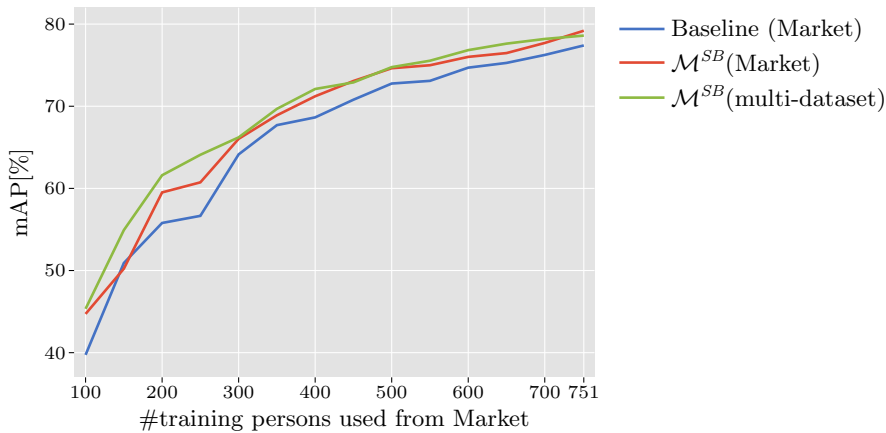
\begin{figure}[t]
    \centering
    \begin{tikzpicture}
        \begin{axis}[
            inner sep=0pt,outer sep=0pt,
            ylabel style={yshift=-15pt},
            xlabel style={yshift=3pt},
            yticklabel style = {font=\scriptsize,xshift=-0.3ex},
            xticklabel style = {font=\scriptsize,yshift=-0.3ex},
            tick label style={/pgf/number format/assume math mode=true},
            tick align=outside,
            major tick length=2pt,
            every tick/.style={black, thin},
            tick pos=left,
            axis line style={draw=none},
            axis background/.style={fill=black!11!white},
            grid=both,
            grid style={white},
            xtick=        {2,   4,   6,   8,  10,  12,  14, 15.02},
            xticklabels={100, 200, 300, 400, 500, 600, 700,   751},
            xlabel = {\#training persons used from Market},
            xmin=1.7, xmax=15.3,
            ylabel = {mAP[\%]},
            ytick = {0.0, 0.1, 0.2, 0.3, 0.4, 0.5, 0.6, 0.7, 0.8, 0.9, 1},
            yticklabels = {0,10, 20, 30, 40, 50, 60, 70, 80, 90, 100},
            ymin=0.37,ymax=0.82,
            legend cell align={left},
            height=6.5cm,
            legend style={draw=none, cells={align=left}, legend pos=outer north east},
            width=8.5cm, 
        ]
        \legend{Baseline (Market), \singlebranch (Market), \singlebranch (multi-dataset)}
        \addplot[better_blue, line width = 1pt]
        coordinates {(2, 0.39735991) (3, 0.50915450) (4, 0.55789896) (5, 0.56656882) (6, 0.64138130) (7, 0.67698925) (8, 0.68643371) (9, 0.7079359) (10, 0.72762542) (11, 0.73089905) (12, 0.74695586) (13, 0.7527334) (14, 0.76243628) (15.02, 0.774)};\label{plot:lessdata};

        \addplot[better_red, line width = 1pt]
        coordinates {(2, 0.44719884) (3, 0.50220909) (4, 0.59509648) (5, 0.60735810) (6, 0.66032272) (7, 0.68896652) (8, 0.71212874) (9, 0.73072528) (10, 0.74626304) (11, 0.74990794) (12, 0.76008895) (13, 0.76472069) (14, 0.77702509) (15.02, 0.7919)};\label{plot:lessdata_multitask};

        \addplot[better_green, line width = 1pt]
        coordinates {(2, 0.4534) (3, 0.5492) (4, 0.6160) (5, 0.6409) (6, 0.6621) (7, 0.6968) (8, 0.7210) (9, 0.7290) (10, 0.7476) (11, 0.7554) (12, 0.7683) (13, 0.7762) (14, 0.7819) (15.02, 0.7860)};\label{plot:lessdata_multitask_multidataset};
        \end{axis}
    \end{tikzpicture}
    \caption{
        Learning curves for the ReID performance.
        MTL (\ref{plot:lessdata_multitask}) consistently outperforms the triplet ReID baseline (\ref{plot:lessdata}).
        Joint multi-dataset learning with MPII and LIP (\ref{plot:lessdata_multitask_multidataset}) gives an additional, albeit smaller, boost.
    }
    \label{fig:reduced_data}
\end{figure}

\subsection{Effect of the Training Set Size}
In Figure~\ref{fig:reduced_data} we analyze the benefits provided by multi-task and multi-dataset learning over the baseline as a function of the amount of ReID training data used.
These learning curves show that the improvements hold across a wide range of training set sizes and suggests that the combination of diverse sources of supervision will remain a relevant topic even as available datasets grow in size.

\subsection{Qualitative Results}
\begin{figure}[!h]
    \centering%
    \begin{subfigure}{0.152\textwidth}%
        \includegraphics[width=\textwidth, height=0.23\textheight]{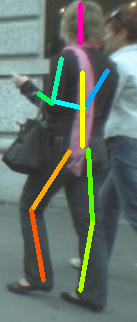}%
    \end{subfigure}~%
    \begin{subfigure}{0.31\textwidth}%
        \includegraphics[width=\textwidth, height=0.23\textheight]{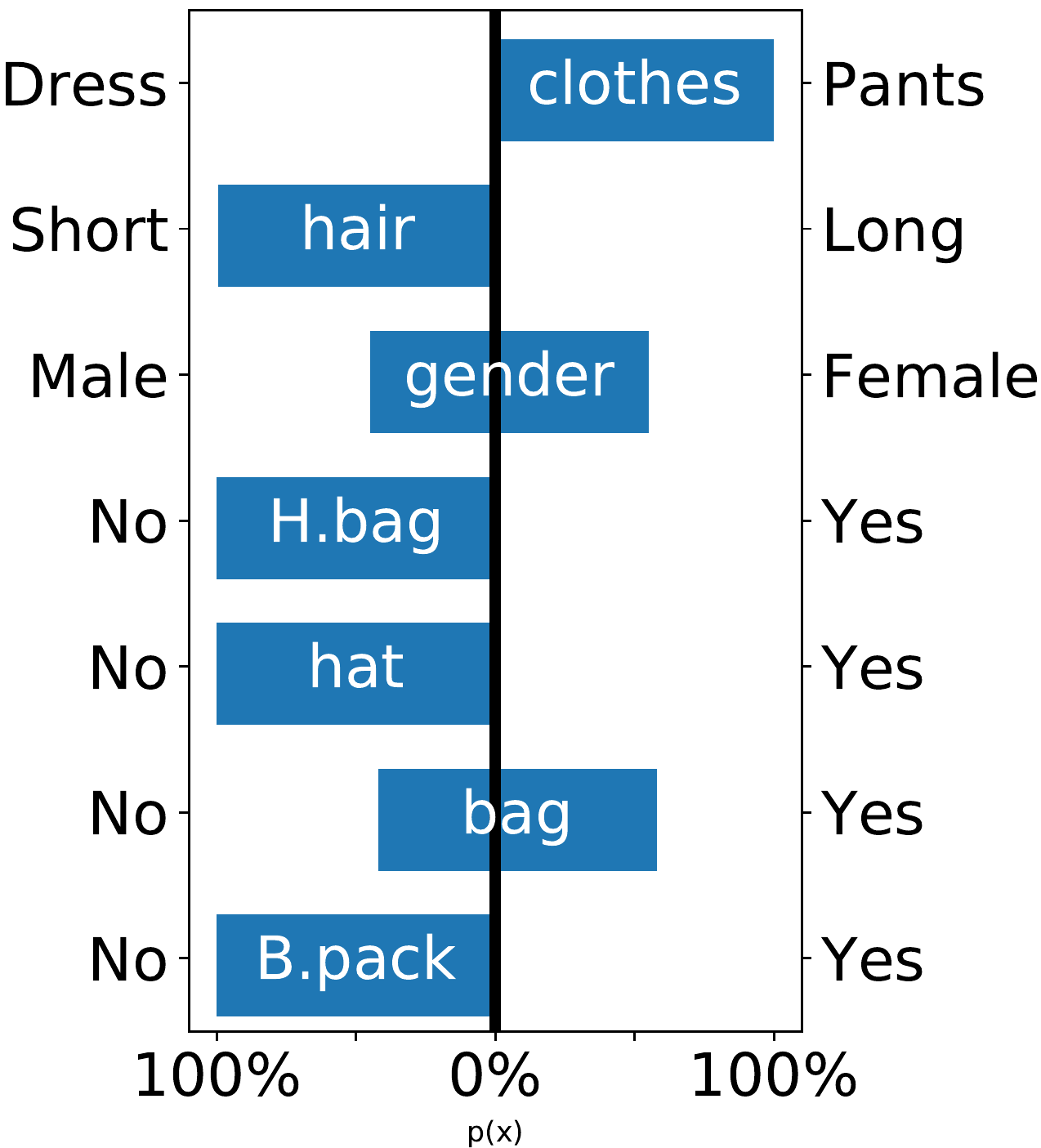}%
    \end{subfigure}~~~~%
    \begin{subfigure}{0.152\textwidth}%
        \includegraphics[width=\textwidth, height=0.23\textheight]{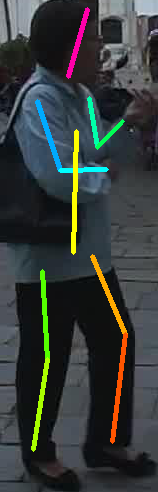}%
    \end{subfigure}~%
    \begin{subfigure}{0.31\textwidth}%
        \includegraphics[width=\textwidth, height=0.23\textheight]{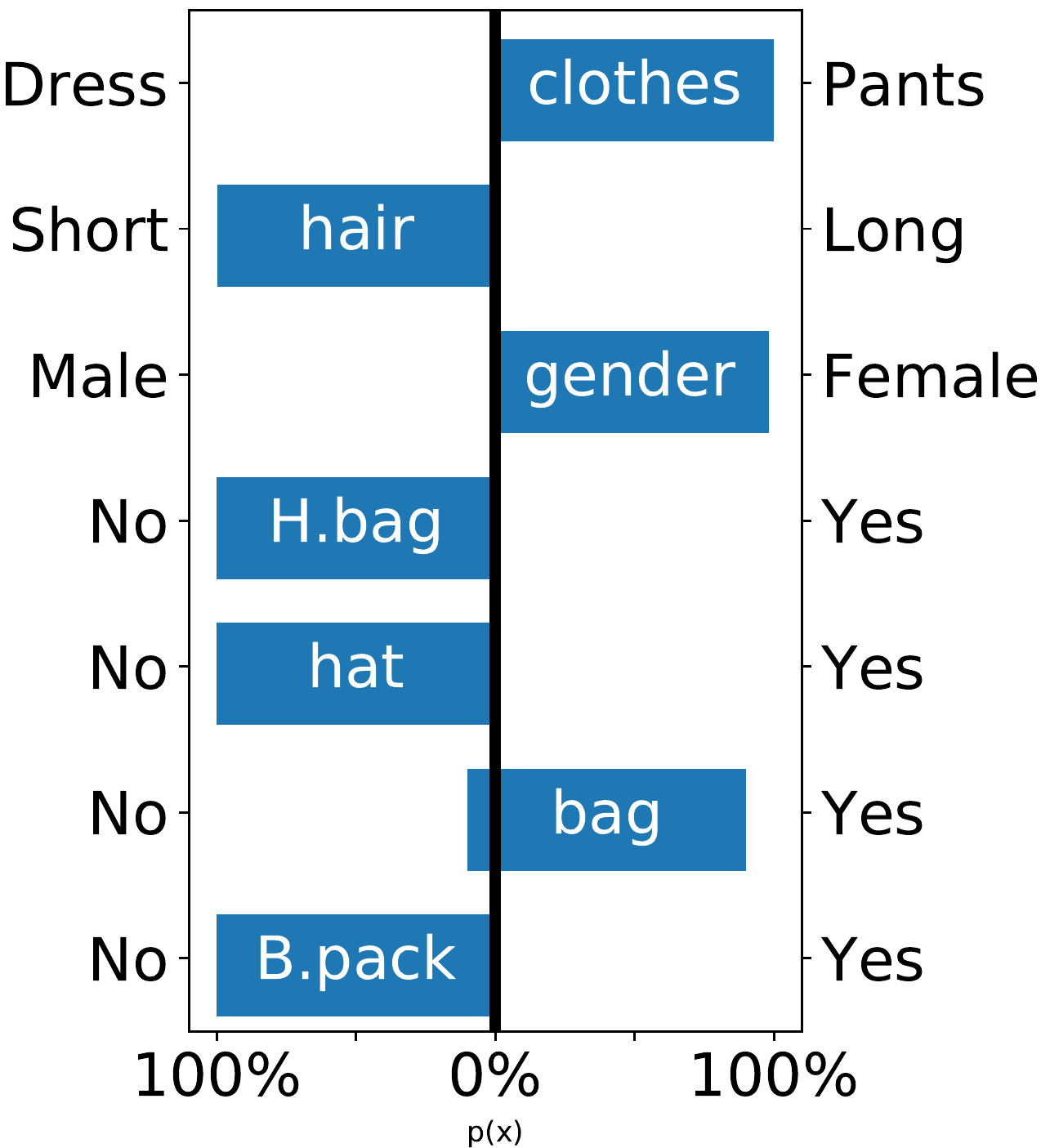}%
    \end{subfigure}%

    \begin{subfigure}{0.152\textwidth}%
        \includegraphics[width=\textwidth, height=0.23\textheight]{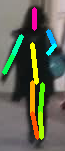}%
    \end{subfigure}~%
    \begin{subfigure}{0.31\textwidth}%
        \includegraphics[width=\textwidth, height=0.23\textheight]{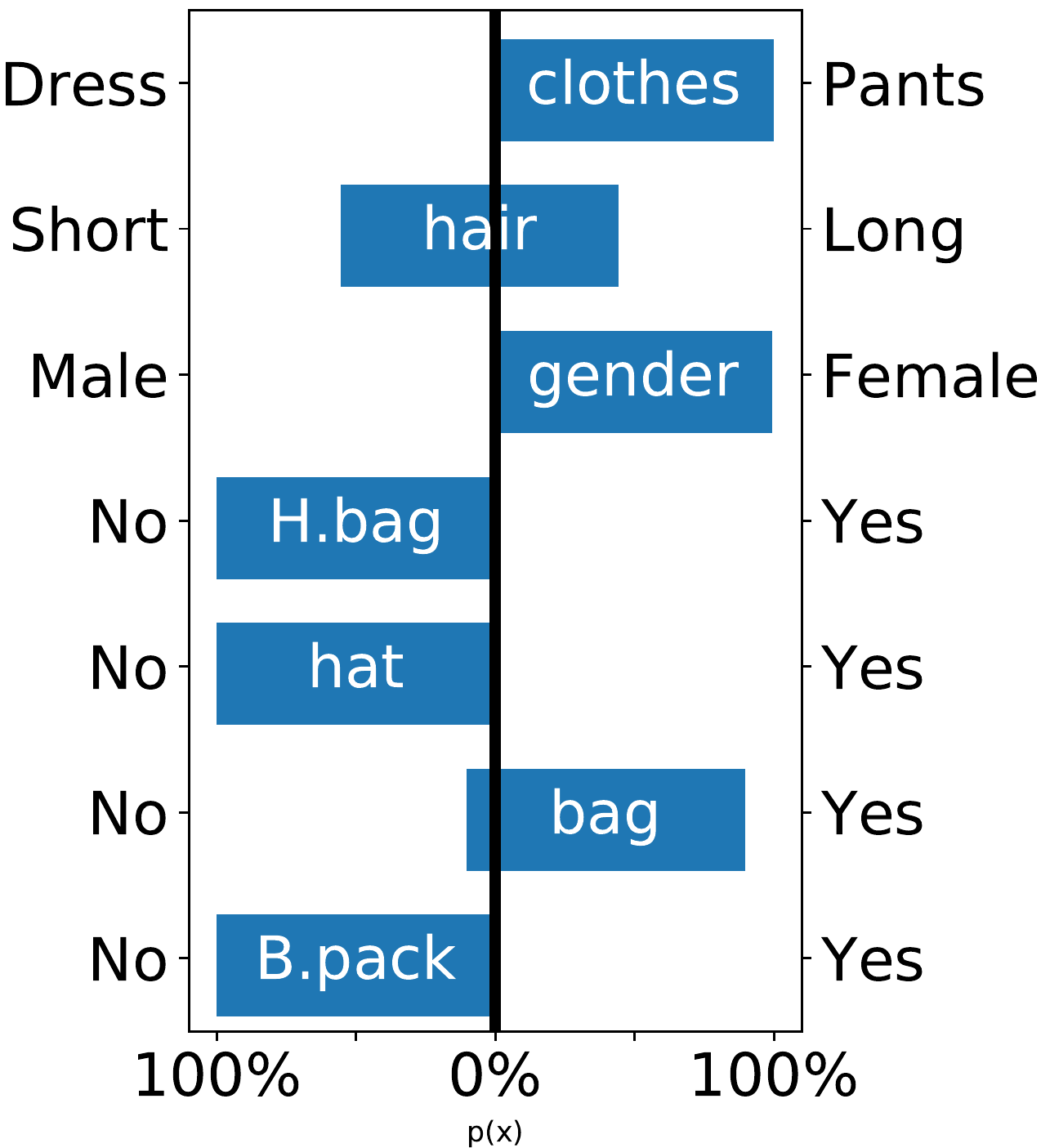}%
    \end{subfigure}~~~~%
    \begin{subfigure}{0.152\textwidth}%
        \includegraphics[width=\textwidth, height=0.23\textheight]{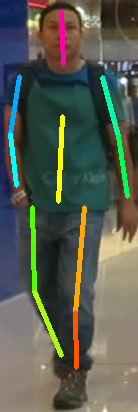}%
    \end{subfigure}~%
    \begin{subfigure}{0.31\textwidth}%
        \includegraphics[width=\textwidth, height=0.23\textheight]{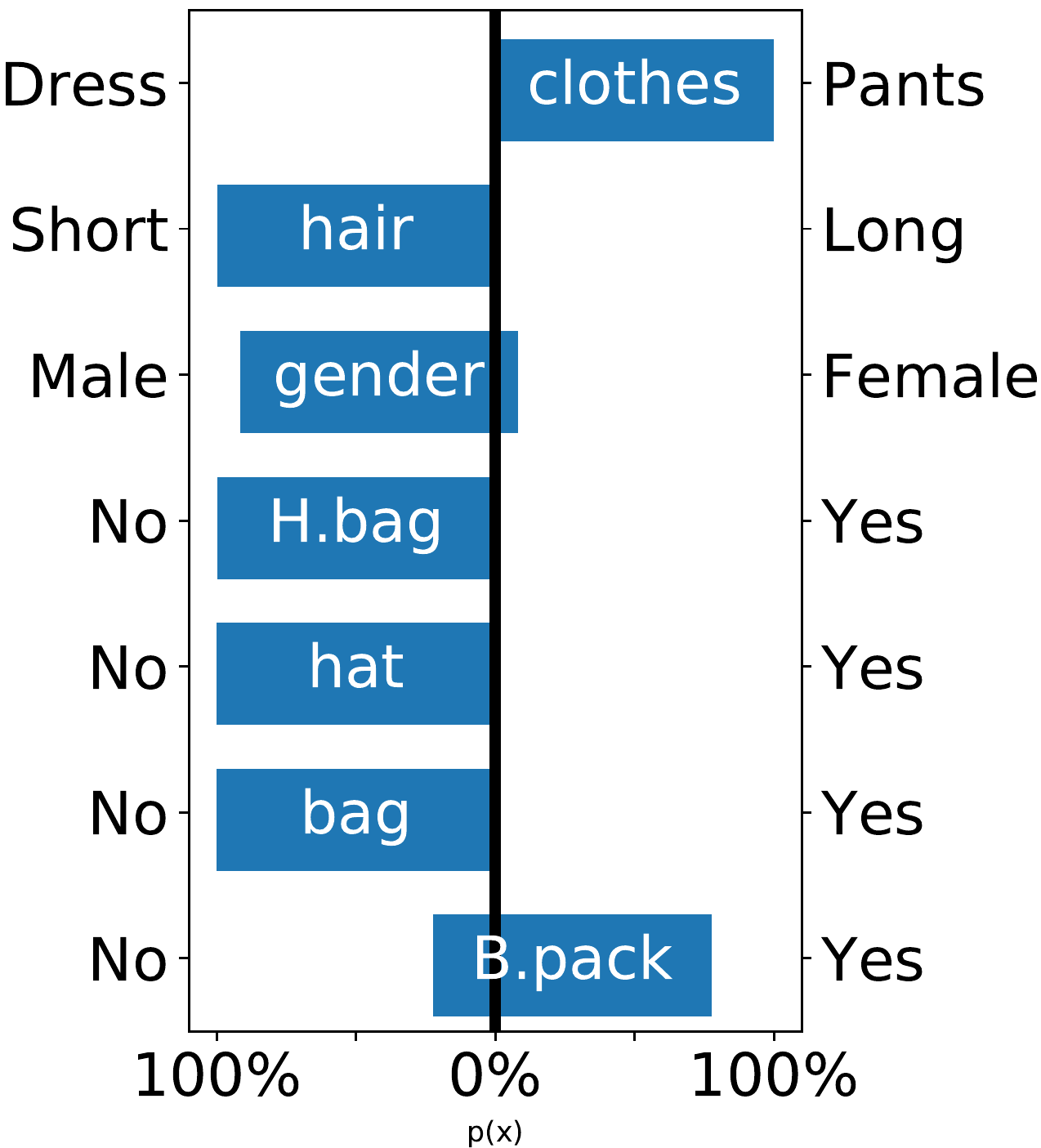}%
    \end{subfigure}%

    \begin{subfigure}{0.152\textwidth}%
         \includegraphics[width=\textwidth]{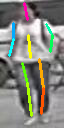}%
    \end{subfigure}~%
     \begin{subfigure}{0.31\textwidth}%
         \includegraphics[width=\textwidth]{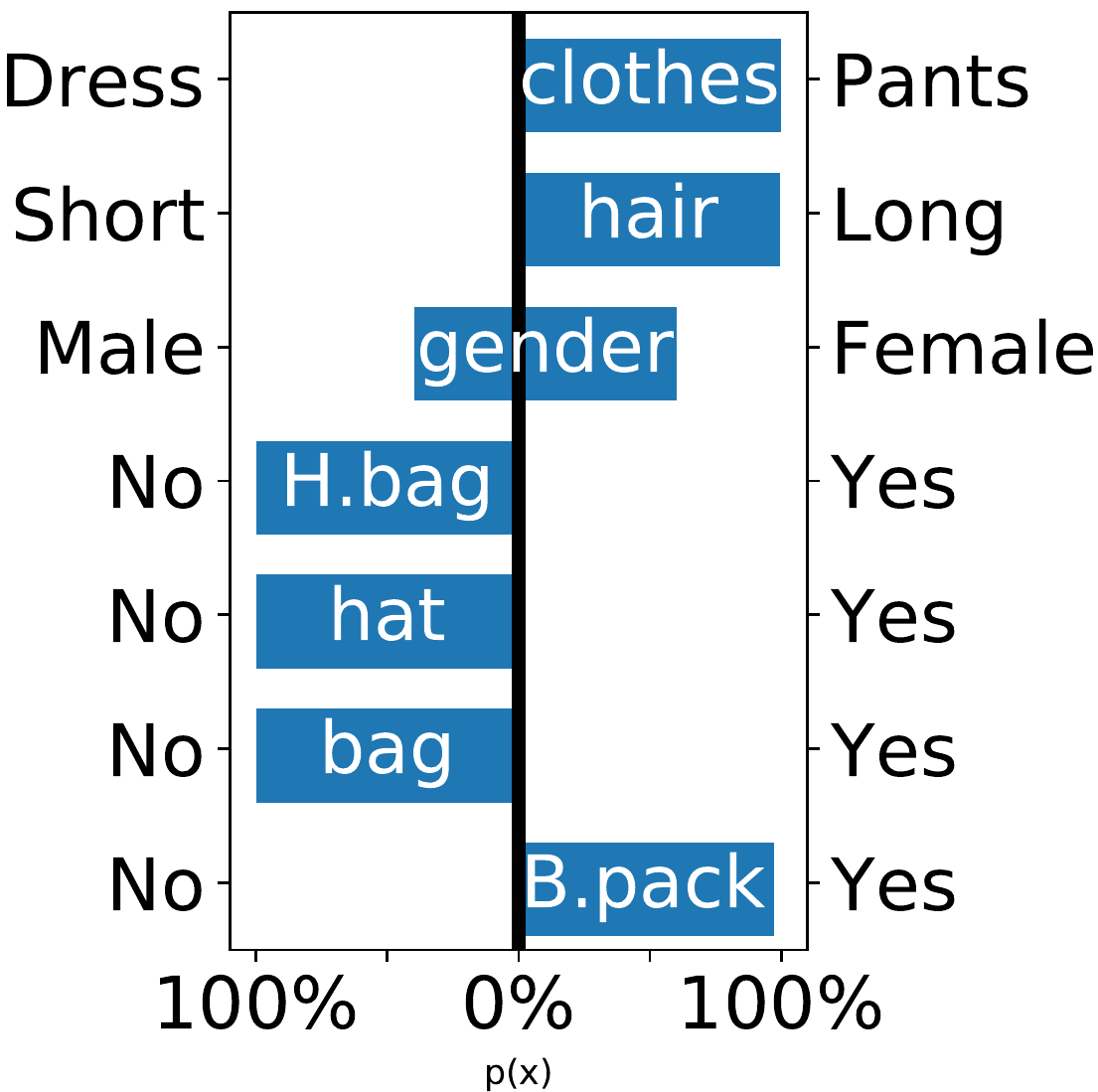}%
     \end{subfigure}~~~~%
     \begin{subfigure}{0.152\textwidth}%
         \includegraphics[width=\textwidth]{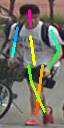}%
     \end{subfigure}~%
     \begin{subfigure}{0.31\textwidth}%
         \includegraphics[width=\textwidth]{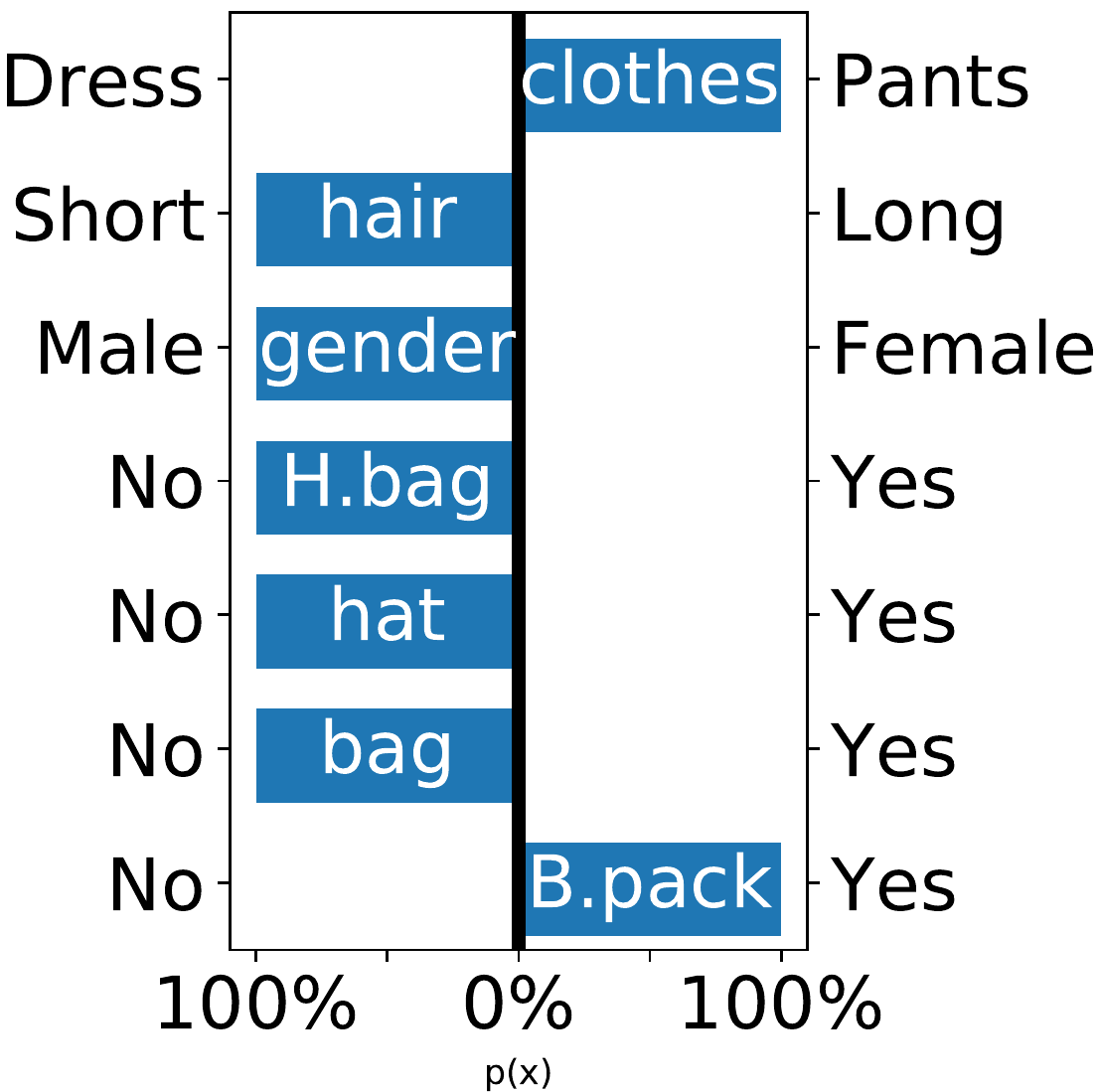}%
     \end{subfigure}%

     \begin{subfigure}{0.152\textwidth}%
         \includegraphics[width=\textwidth]{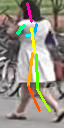}%
     \end{subfigure}~%
     \begin{subfigure}{0.31\textwidth}%
         \includegraphics[width=\textwidth]{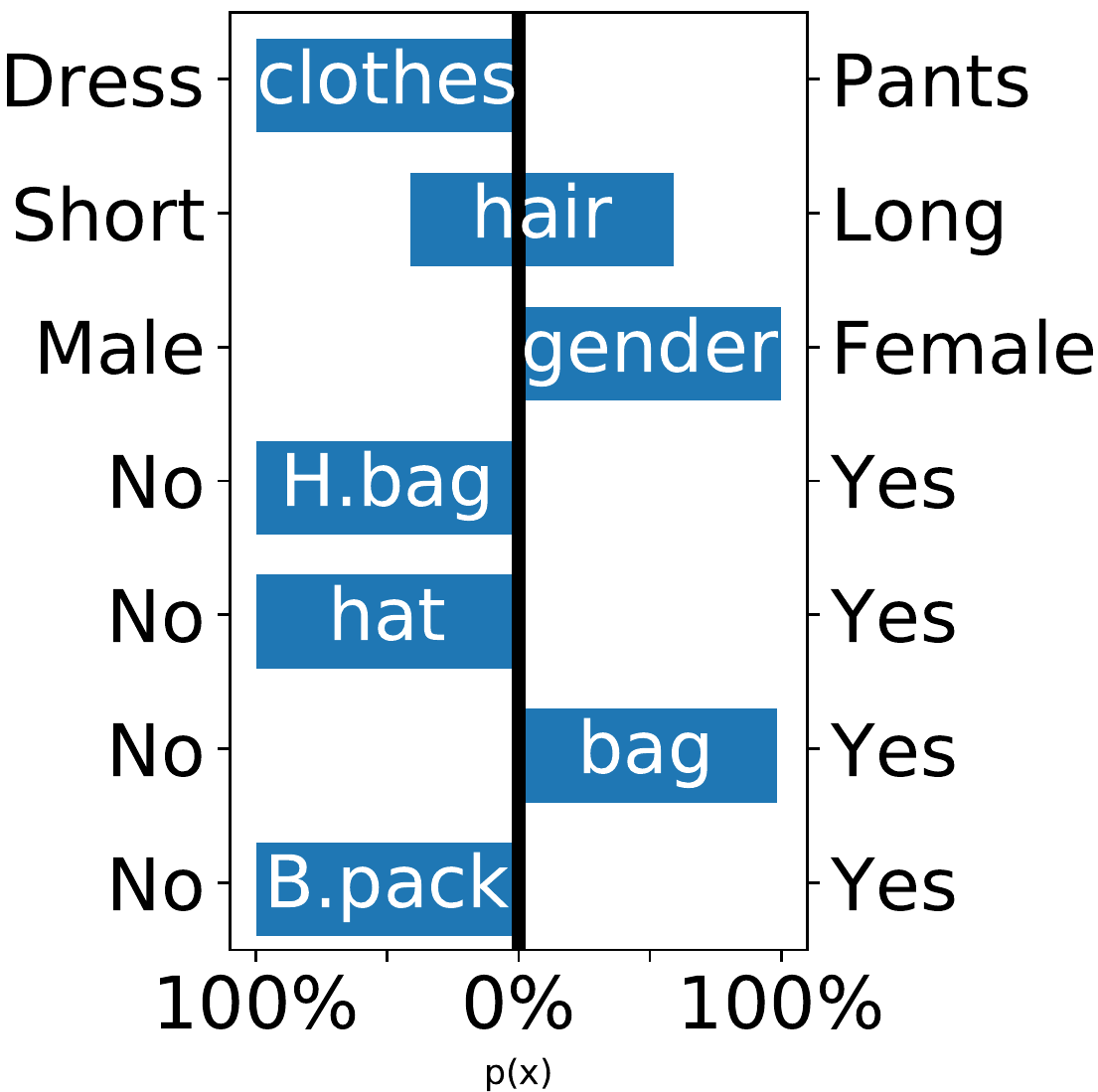}%
     \end{subfigure}~~~~%
     \begin{subfigure}{0.152\textwidth}%
         \includegraphics[width=\textwidth]{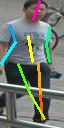}%
     \end{subfigure}~%
     \begin{subfigure}{0.31\textwidth}%
         \includegraphics[width=\textwidth]{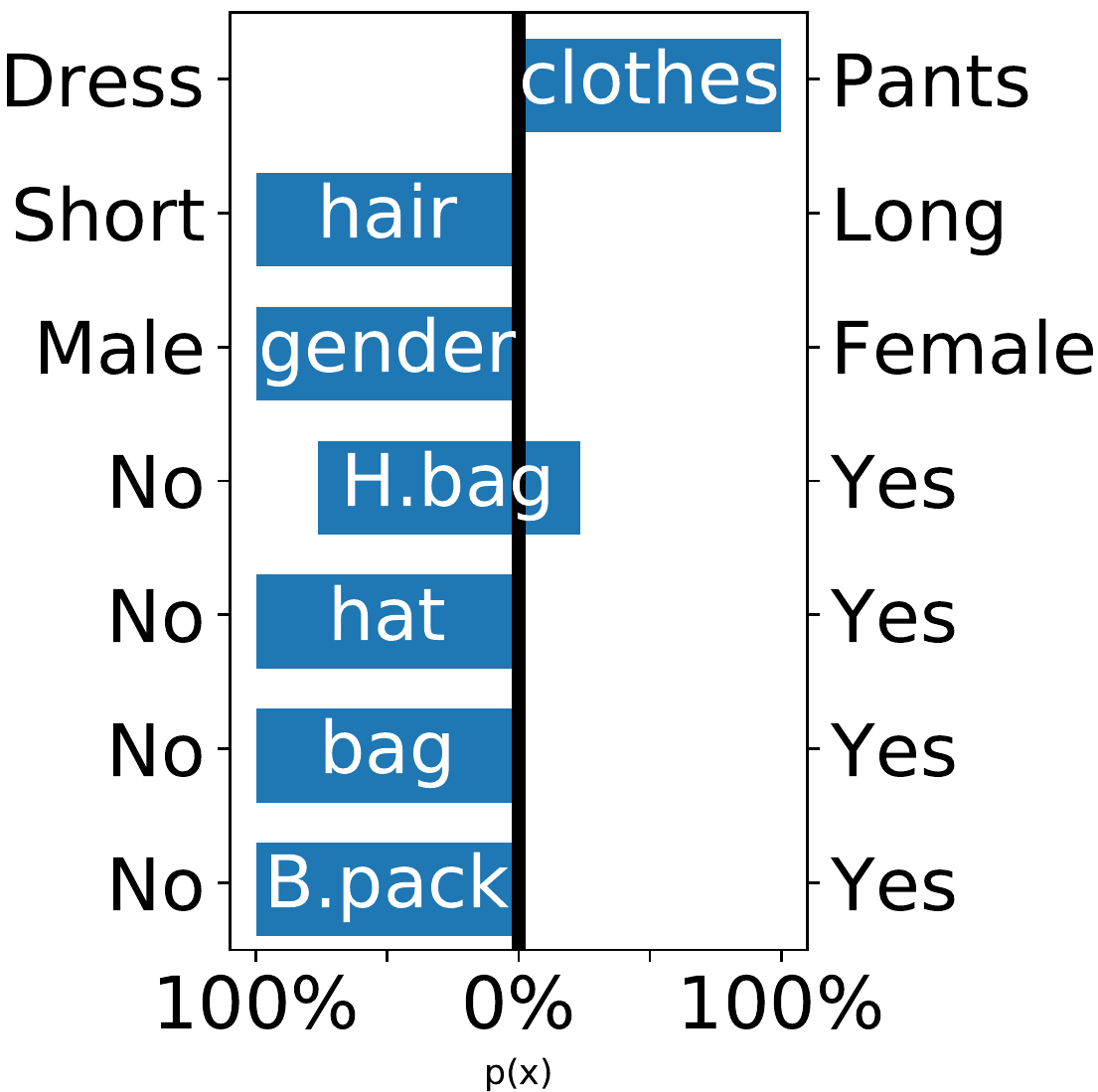}%
     \end{subfigure}%
    \caption{
        Given person detections, we perform pose estimation and attribute classification jointly with person re-identification and part segmentation (not visualized) using a shared CNN backbone with small task-specific heads.
    }
    \label{fig:attributes}
\end{figure}

\begin{figure}[!h]
    \definecolor{mycyan}{HTML}{00A8A8}
    \definecolor{mypink}{HTML}{FF5757}
    \centering
    \begin{subfigure}{0.495\textwidth}%
        \includegraphics[width=\textwidth]{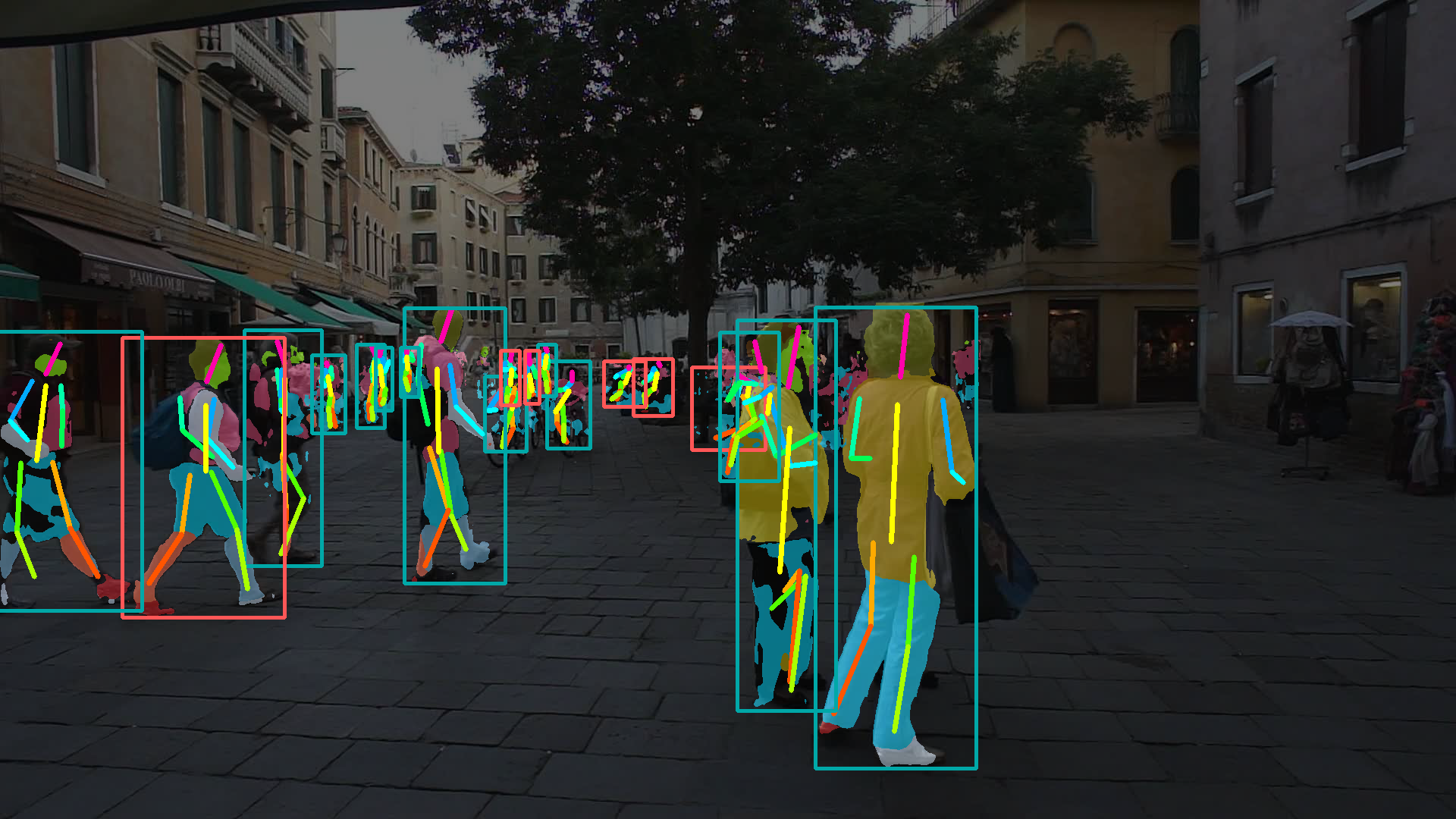}%
    \end{subfigure}\,%
    \begin{subfigure}{0.495\textwidth}%
        \includegraphics[width=\textwidth]{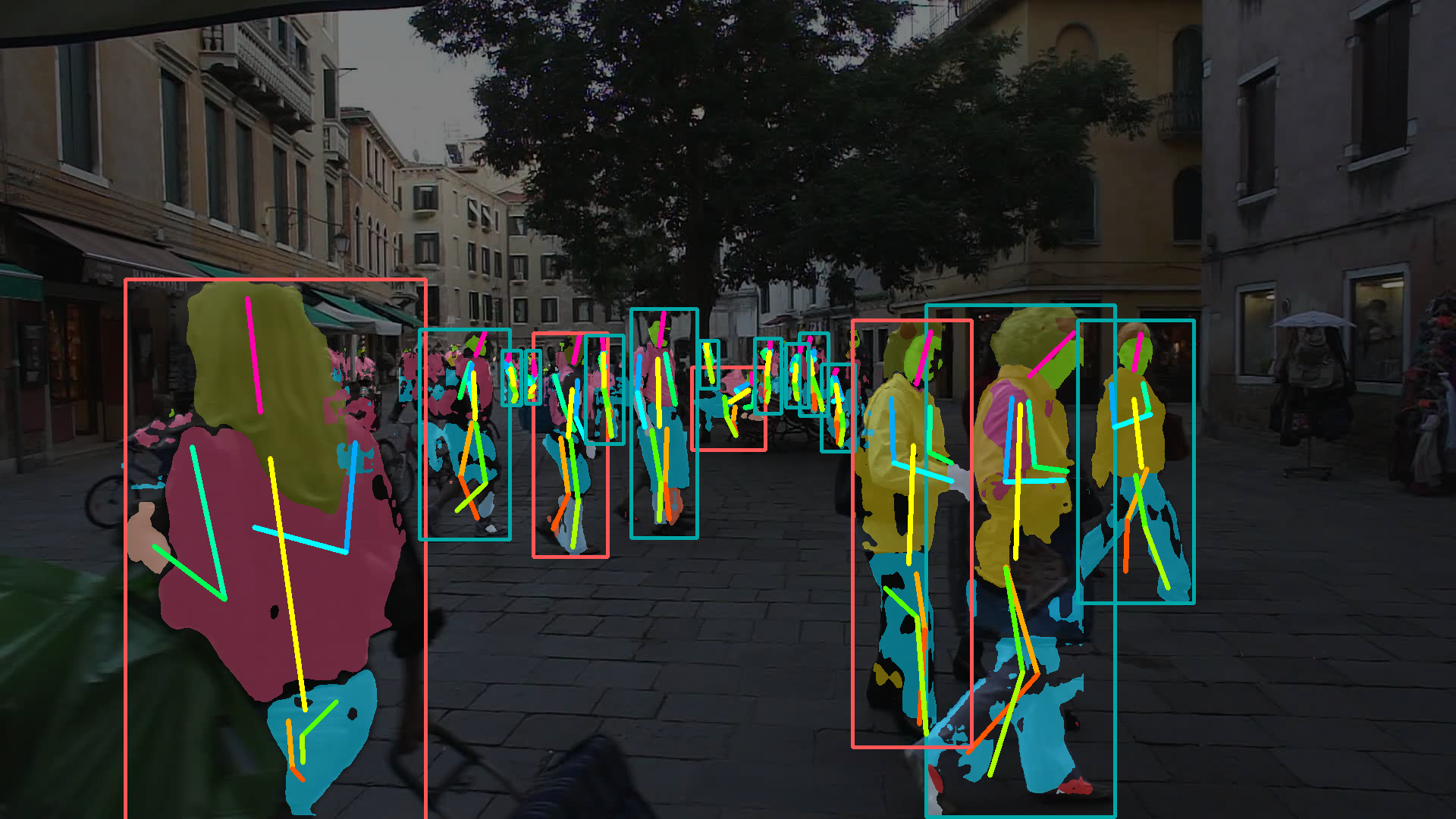}%
    \end{subfigure}%

    \begin{subfigure}{0.495\textwidth}%
        \includegraphics[width=\textwidth]{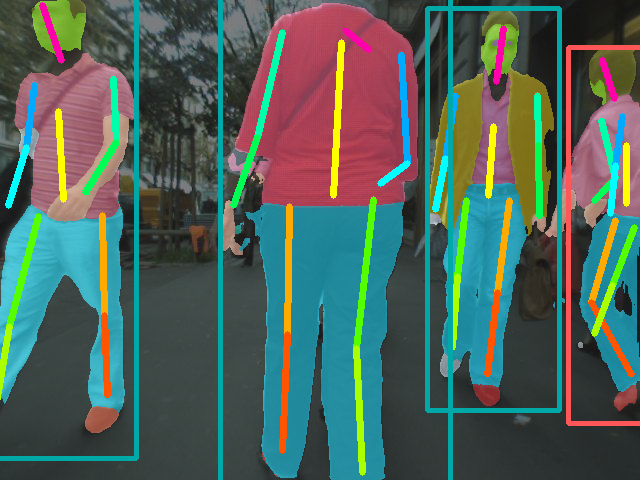}%
    \end{subfigure}\,%
    \begin{subfigure}{0.495\textwidth}%
        \includegraphics[width=\textwidth]{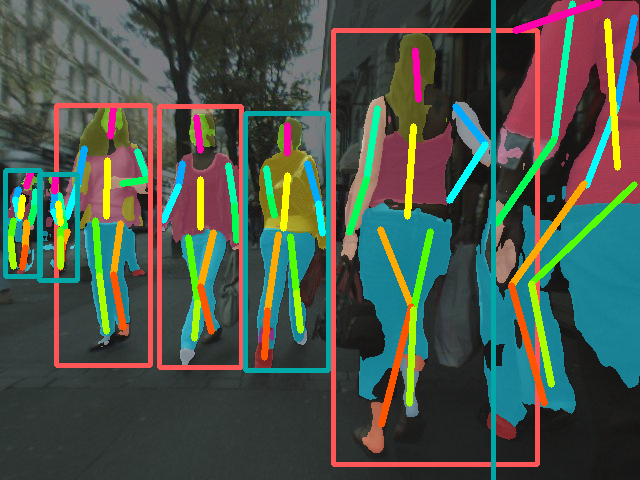}%
    \end{subfigure}%

    \begin{subfigure}{0.495\textwidth}%
        \includegraphics[width=\textwidth]{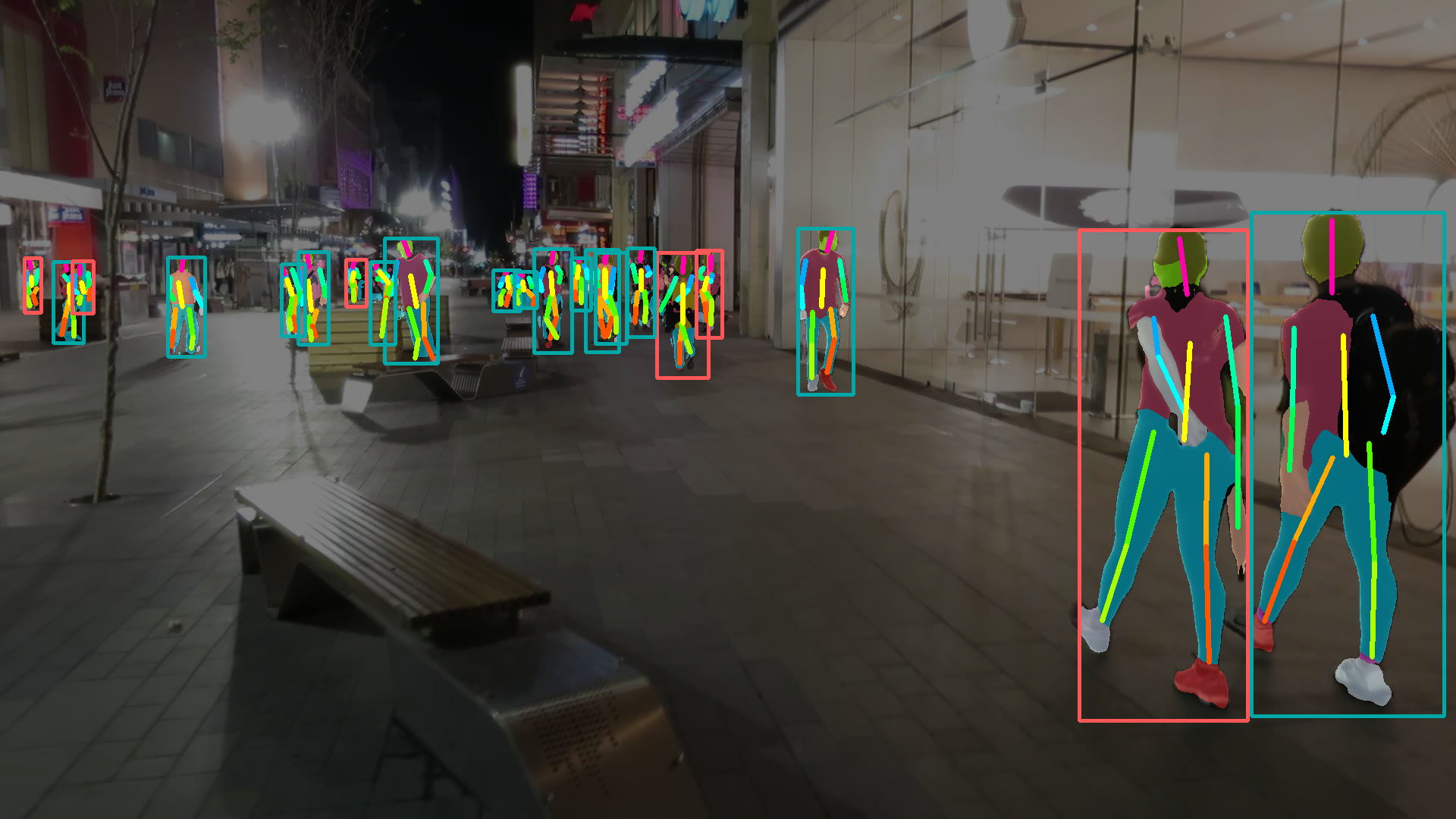}%
    \end{subfigure}\,%
    \begin{subfigure}{0.495\textwidth}%
        \includegraphics[width=\textwidth]{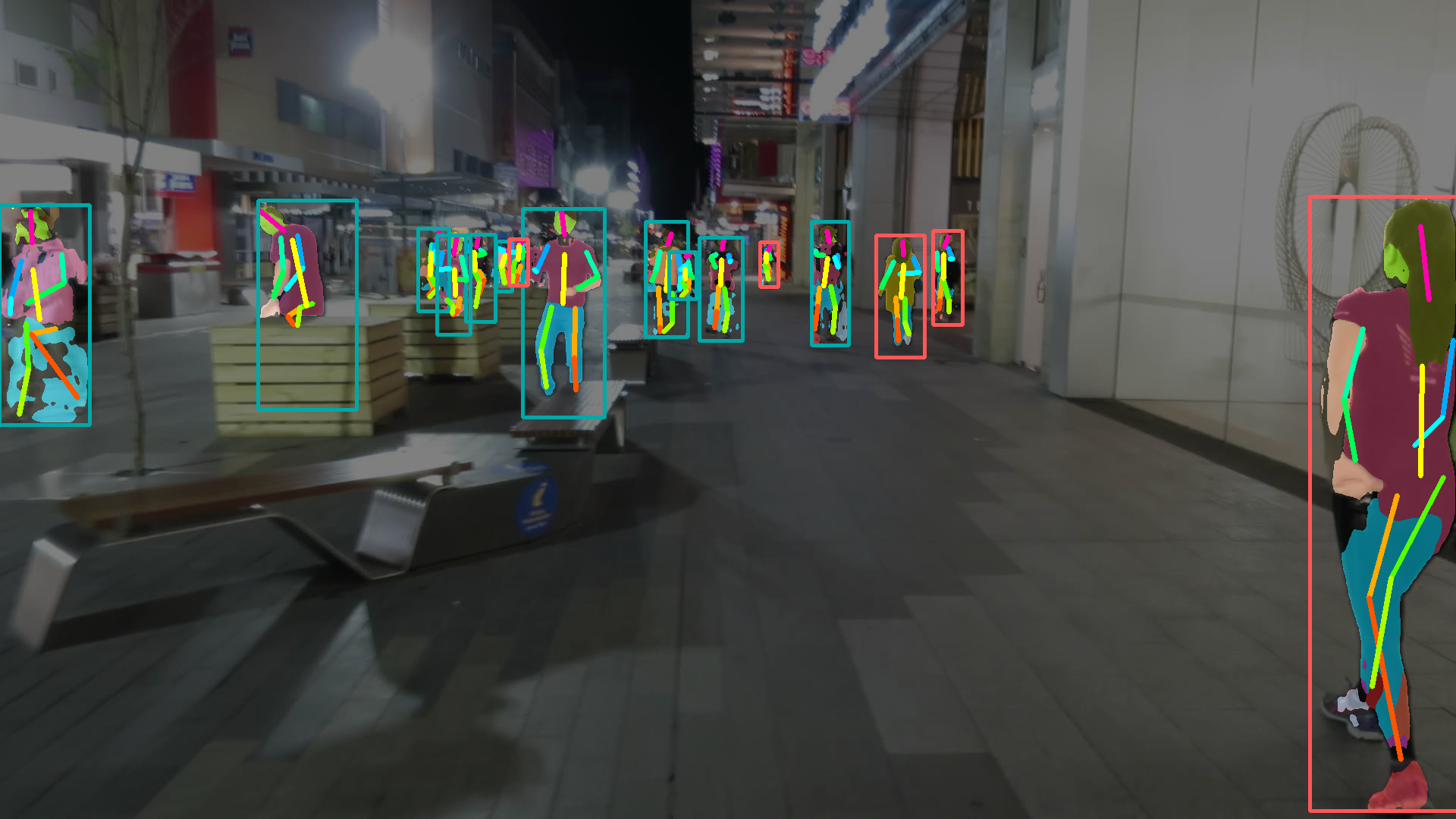}%
    \end{subfigure}%

    \begin{subfigure}{0.495\textwidth}%
        \includegraphics[width=\textwidth]{images/mot_11_0170.png}%
    \end{subfigure}\,%
    \begin{subfigure}{0.495\textwidth}%
        \includegraphics[width=\textwidth]{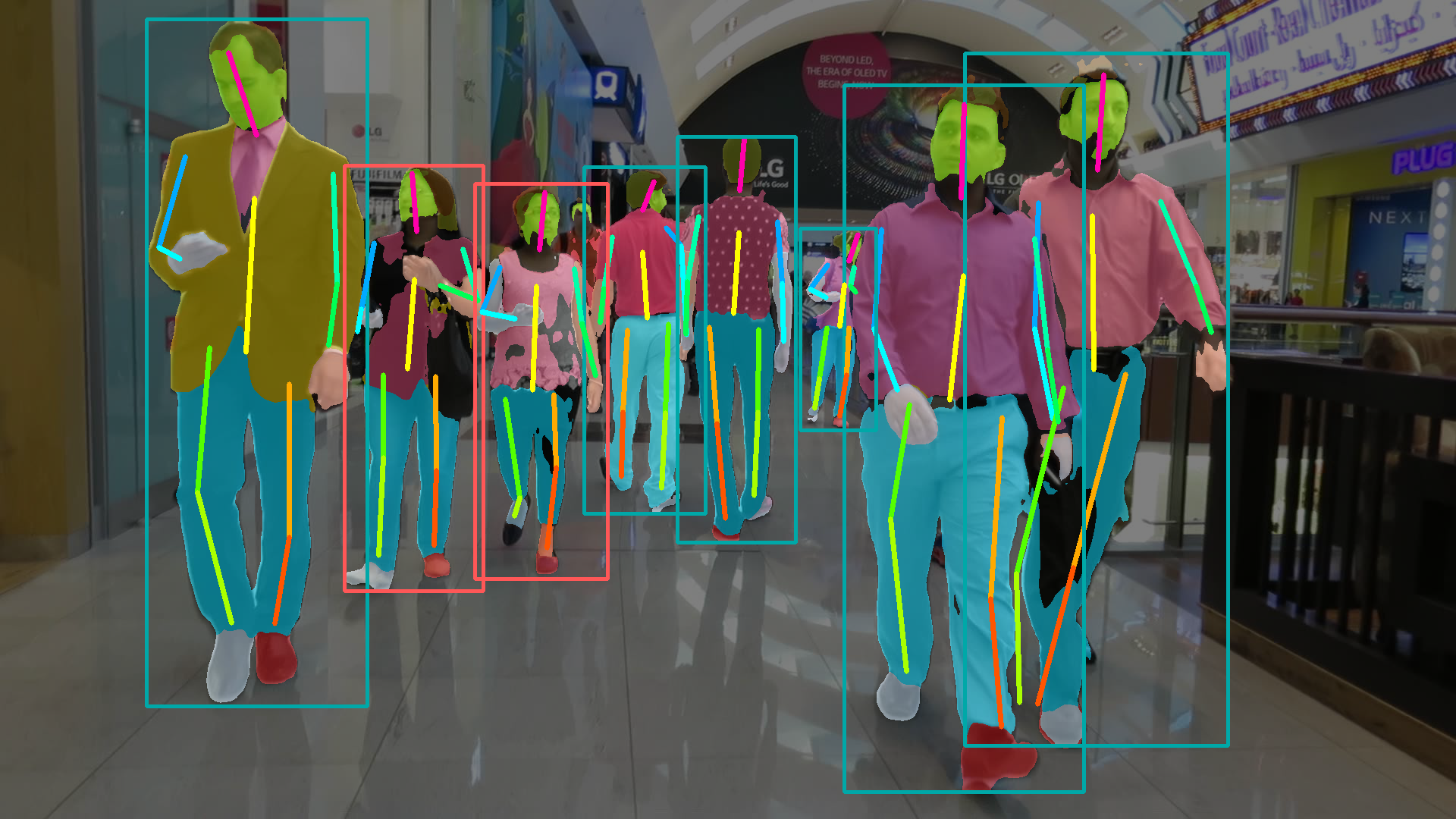}%
    \end{subfigure}%
    \caption{
        Pose estimation and part segmentation results shown on MOT16 sequences~\cite{Milan16Arxiv}.
        We use ground truth bounding boxes for this visualization, but detection boxes can be used instead.
        The bounding box colors correspond to gender predictions ({\color{mypink}female}, {\color{mycyan}male}).
    }
    \label{fig:qualitative_results_mot}
\end{figure}

Figure~\ref{fig:qualitative_mot} shows qualitative results for attribute classification, pose estimation and part segmentation on MOT16 sequences~\cite{Milan16Arxiv}, showing that our network can also generalize to other datasets.
Figure~\ref{fig:attributes} shows additional attribute classification results for binary attributes, both on MOT16 sequences, as well as the Market dataset.
Figure~\ref{fig:qualitative_results_mot} shows several additional frames from MOT16 sequences.
Both pose estimation and part segmentations are shown, as well as the gender prediction visualized by the bounding box color.
Note that both additional attributes are classified and ReID embeddings are generated by the same model, but these are not visualized.
\newpage

\clearpage

\section{Discussion}
\vspace*{-6pt}
We have shown that several visual person understanding task can be tackled using a single unified model.
We saw interesting synergies between different person-related tasks and datasets.
Focusing on ReID, we were able to improve the mAP score by 1.7\% by either utilizing additional tasks through automatic annotations, or multi-dataset training.
Especially with the latter approach, our unified model can roughly match or even outperform the separately trained task- and dataset-specific baselines.
More parameter sharing in the backbone proved beneficial for most tasks.
For the particular case of ReID and pose estimation, the interference was overcome through a simple split of the network output.

Given our positive results, several directions open up for future work.
Further person-related tasks and datasets can be added into the training.
Better task heads from the specific domains or more complex multi-task training schemes will likely achieve further improvements.
Ultimately, it will be interesting to integrate the tasks directly into a detector as an additional person understanding head.
Combinations such as detection and ReID~\cite{Xiao17CVPR}, or instance segmentation and pose estimation~\cite{He17CVPR} have been evaluated, however there are no detectors jointly tackling all the tasks.

Most of the computation in our best-performing model is shared across tasks, adding almost no overhead to the single-task baseline.
On a GTX 1080 Ti GPU it processes 50 person crops per second without batching and 122 with a batch size of 10.
This could be further improved with a lighter backbone, rendering it well-suited for robotics applications, where resource constraints typically prohibit running several models in parallel.
\vspace*{-2pt}

\section{Conclusion}
\vspace*{-6pt}
We have shown that is possible to train a single model for four important person-centric tasks, offering a holistic person understanding without running separate models in parallel.
We evaluated how different backbones and normalizations affect the resulting performance and found that GroupNorm is crucial for multi-dataset training.
While person ReID and pose estimation interfere to a certain degree, a simple split of the backbone output can largely resolve this issue.
Even though we make no complex modification to our ResNet backbone and use simple task heads, we can exploit task synergies using multi-task learning.
Both automatic annotations and multiple datasets can be used to improve results and especially in the latter case we can roughly match or even outperform the single-task baseline scores.
Based on these results, we see several interesting research directions to further advance visual person understanding.
Nevertheless, even our current multi-task model is highly relevant for applications where people and machines interact, especially when computational resources are limited.

\PAR{Acknowledgents.}
This project was funded, in parts, by ERC Consolidator Grant project ``DeeViSe'' (ERC-CoG-2017-773161) and the BMBF projects ``FRAME'' (16SV7830) and ``PARIS'' (16ES0602).
Istvan Sarandi's research is funded by a grant from the Bosch Research Foundation.
Most experiments were performed on the RWTH Aachen University CLAIX 2018 GPU Cluster.

\bibliographystyle{splncs03}
\bibliography{abbrev_short,bib}

\end{document}